%% file: meta_gr_attack.tex
\PassOptionsToPackage{table}{xcolor}
\PassOptionsToPackage{provide*=*}{babel}
\documentclass{fairmeta}

\usepackage[utf8]{inputenc}

\usepackage{adjustbox}
\usepackage{wrapfig}
\usepackage{needspace}
\usepackage[inline]{enumitem}
\usepackage{amssymb}
\usepackage{mathtools}
\renewcommand{\authorformat}[2][]{\mbox{\sffamily\bfseries #2$^{#1}$}}

\providecommand{\Description}[1]{}
\input{presentation/table_style}
\colorlet{rowgray}{metabg}
\definecolor{headergray}{RGB}{220,230,250}

\newcommand{\methodname}{\mbox{AlignPoison}}

\title{\mbox{AlignPoison: Conditional Targeted}
\mbox{Poisoning of Generative Recommenders}
\mbox{via Decoding-Aligned Prefixes}}

\author[1,*]{Zhuoxuan Zhang}
\author[1,*]{Jinhao Duan}
\author[1]{Xinyu Zhao}
\author[2]{Jinduo Guo}
\author[3]{Shuai Shao}
\author[3]{Rui Li}
\author[3]{Luke Simon}
\author[3]{Zhun Deng}
\author[1]{Tianlong Chen}
\author[3]{Bo Sun}

\affiliation[1]{University of North Carolina at Chapel Hill}
\affiliation[2]{University of Notre Dame}
\affiliation[3]{Meta}
\contribution[*]{Equal contribution.}

\abstract{\input{sections/abstract}}
\date{October 10, 2026}
\hypersetup{
  pdftitle={AlignPoison: Conditional Targeted Poisoning of Generative Recommenders via Decoding-Aligned Prefixes},
  pdfauthor={Zhuoxuan Zhang, Jinhao Duan, Xinyu Zhao, Jinduo Guo, Shuai Shao, Rui Li, Luke Simon, Zhun Deng, Tianlong Chen, Bo Sun}
}

\begin{document}
\maketitle

\input{sections/introduction}
\input{sections/related_work}
\input{sections/preliminaries}
\input{sections/method}
\input{sections/experiments}

\input{sections/conclusion}
\input{sections/acknowledgements}
\input{sections/limitations}
\input{sections/submission_statements}

\clearpage
\bibliographystyle{assets/plainnat}
\bibliography{references}

\clearpage
\beginappendix
\input{sections/appendix}

\end{document}

%% file: presentation/table_style.tex
\usepackage{siunitx}
\definecolor{PTHeader}{HTML}{F3F5F7}
\definecolor{PTHighlight}{HTML}{FDEDE9}
\definecolor{PTInk}{HTML}{27313D}
\definecolor{PTGain}{HTML}{FF0000}
\definecolor{PTRule}{HTML}{4B5563}
\definecolor{PTGroupRule}{HTML}{D5DAE0}
\newcommand{\PaperTableStyle}{%
  \fontsize{9}{10.6}\selectfont\color{PTInk}%
  \setlength{\tabcolsep}{4pt}%
  \renewcommand{\arraystretch}{1.10}%
  \sisetup{detect-weight=true,detect-family=true,group-digits=false}%
  \setlength{\aboverulesep}{0pt}\setlength{\belowrulesep}{0pt}%
  \arrayrulecolor{PTRule}%
}
\newcommand{\PN}[1]{\tablenum[table-format=2.1]{#1}}
\newcommand{\PB}[1]{\textbf{\PN{#1}}}
\newcommand{\PS}[1]{\underline{\PN{#1}}}
\newcommand{\PF}[1]{\tablenum[table-format=1.2]{#1}}
\newcommand{\PG}[1]{\ensuremath{_{\color{PTGain}\uparrow (#1)}}}

%% file: sections/abstract.tex
Generative recommendation models based on autoregressive decoding have been adopted as alternatives to classical ranking-based systems. While their modeling flexibility has been widely studied, their resilience to data poisoning remains largely unexplored. In this work, we ask: how susceptible are generative recommenders to conditional targeted poisoning? Under a realistic setting where an adversary can inject limited malicious interactions and query a recommendation API as a black box, our empirical study shows that generative recommenders are vulnerable: injecting fewer than 1\% synthetic interactions significantly increases the likelihood of promoting malicious targets under the trigger. To characterize this vulnerability, we analyze the synthesis of malicious interactions from two perspectives: (i) decoding alignment and (ii) prefix popularity. We observe that short, context-aligned prefixes that are both common under the model's induced transition dynamics and strongly linked to a frequent trigger consistently amplify targeted promotion. Motivated by this analysis, we develop \methodname{}, a black-box framework that discovers and samples context-aligned prefixes from API outputs. We conduct experiments on benchmarks in unimodal and multimodal generative recommendation settings. Results show that these prefixes outperform competitive baselines, improving ASR lift by up to 16.3 percentage points and highlighting a decoding-dependent, context-sensitive poisoning vulnerability in generative recommendation models.

%% file: sections/introduction.tex
\section{Introduction}
\label{sec:intro}

\begin{wrapfigure}{r}{0.44\linewidth}
  \centering
  \includegraphics[width=\linewidth]{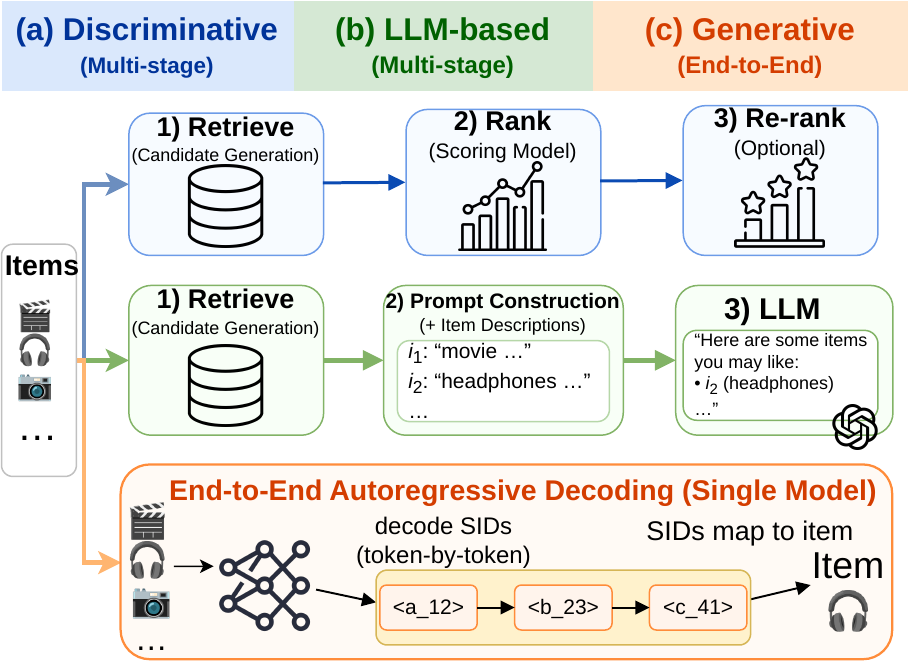}
  \caption{Conventional ((a)--(b)) vs. generative ((c)) recommendation architectures.}
  \label{fig:rec_distinct}
\end{wrapfigure}

The recommendation paradigm is undergoing a fundamental shift from multi-stage ranking pipelines toward end-to-end unified generative architectures~\citep{zhao2026generativesurvey}. By combining user histories and multimodal signals into a single generative task, these systems consistently outperform conventional multi-stage methods \citep{rajput2023recommendersystemsgenerativeretrieval, zhou2025onerectechnicalreport,zhou2025onerecv2}. This strong empirical performance has accelerated their transition from academic prototypes to large-scale industrial deployment, exemplified by systems such as OneRec on major short-video platforms \citep{zhou2025onerectechnicalreport,liu2026onerecthink}. At the same time, this architectural shift introduces a different attack surface: rather than scoring a fixed candidate set or relying on natural-language item descriptions, generative recommenders autoregressively decode semantic item IDs, as illustrated in Figure~\ref{fig:rec_distinct}. Notably, this distinction raises an immediate robustness question. Despite their rapid adoption, it remains unclear whether this shift inherits, mitigates, or amplifies the data poisoning risks documented in traditional sequential systems \citep{Song2020PoisonRecAA,zhang2024lorec,zhao2025diversity}.

Poisoning attacks manipulate exposure and downstream user behavior, threatening platform integrity, safety, and fairness~\citep{Mobasher2007TowardTR,fang2018poisoning,fang2020influence,Song2020PoisonRecAA}. By injecting a small number of crafted user--item interactions into training data, attackers steer recommendations toward a target item in specific user-history contexts~\citep{tang2020revisiting,fang2020influence,Song2020PoisonRecAA,nguyen2023gspattack,ning2026badrec}.
We study conditional targeted poisoning of end-to-end generative recommenders with limited interaction injection and black-box API access.

\noindent \textbf{RQ1 (Susceptibility)}: Are end-to-end generative recommenders vulnerable to conditional targeted poisoning at small poisoning ratios? Even simple trigger-conditioned injections substantially increase target exposure (Section~\ref{sec:warm-up_study}).

\noindent \textbf{RQ2 (Mechanism)}: What properties of synthetic interaction histories (malicious prefixes) determine attack effectiveness in generative recommenders? We identify two key properties: \emph{decoding alignment} and \emph{prefix popularity} (Section~\ref{sec:method}).

Based on these insights, we propose \methodname{}, which discovers context-aligned prefixes through API queries and injects synthetic histories to reinforce trigger$\rightarrow$target associations. Across two Amazon product datasets and a multimodal micro-video dataset, \methodname{} outperforms poisoning baselines while preserving benign recommendation quality.

To our knowledge, attacks of end-to-end generative recommenders remains under-explored. Prior studies primarily address ranking-based, representation-based, or LLM-based recommenders~\citep{Mobasher2007TowardTR,fang2018poisoning,fang2020influence,Song2020PoisonRecAA,tang2020revisiting,nguyen2023gspattack,Ning_2024,wang2024clear}. These systems score candidates, match embeddings, or use natural-language inputs and item descriptions~\citep{wang2025idfree}. Our setting instead concerns direct decoding of structured item tokens, motivating a decoding-aware study of how malicious histories affect generation.
Our contributions are:
\begin{itemize}[leftmargin=*]
  \item \textbf{Empirical vulnerability study:} We demonstrate conditional targeted promotion using simple synthetic histories at small poisoning ratios (RQ1).
  \item \textbf{Mechanistic analysis:} We identify two key dimensions of malicious prefixes, \emph{decoding alignment} and \emph{prefix popularity}, and characterize how their interaction drives attack success (RQ2).
  \item \textbf{Black-box framework:} We develop a black-box prefix discovery, optimization, and adaptive sampling framework that recovers short, context-aligned prefixes using only API outputs.
  \item \textbf{Extensive evaluation:} We validate on two Amazon Review subsets (\emph{Office\_Products} and \emph{Industrial\_and\_Scientific}) and the micro-video dataset MicroLens-100K, with comparisons to strong baselines, defenses, and ablations.
\end{itemize}
\FloatBarrier

%% file: sections/related_work.tex
\section{Related Work}
\label{sec:related}

\subsection{Generative Recommendation}
Generative recommendation reframes retrieval as sequence generation, with models decoding item identifiers rather than matching embeddings. Early work such as TIGER~\citep{rajput2023recommendersystemsgenerativeretrieval} introduced encoder--decoder retrieval with semantic IDs, while subsequent systems extended this paradigm to industrial and multimodal settings through stronger architectures, alternative decoding schemes, and richer tokenization strategies, including OneRec, MTGR, NAR4Rec, MDGR, and related semantic-ID methods~\citep{deng2025onerecunifyingretrieverank,han2025mtgr,ren2024non,shah2025masked,feng2022recommender,wang2024eager,wei2025cofirec,geng2023vip5,luo2025qarm,tan2024idgenrec,Yang2024UnifyingGA,zhou2025onerecv2}. A parallel line integrates language interfaces into recommendation, for example through unified prompting or lightweight LLM adaptation~\citep{geng2022recommendation,zheng2024adapting}. We study conditional item-promotion attacks on these structured-token recommenders.

\subsection{Data Poisoning Attacks in Recommendation and Autoregressive Models}
Prior work explores recommender
poisoning~\citep{Mobasher2007TowardTR,Song2020PoisonRecAA,zhao2025diversity,wang2024clear,li2025sntba},
adversarial prompting and poisoning in language
models~\citep{Zou2023UniversalAT,Wan2023PoisoningLM,souly2025poisoning},
and text or profile manipulation in language-based
recommendation~\citep{wang2025idfree,ning2026badrec,Ning_2024}.
We study semantic-ID generative recommenders under top-$K$ query
access and interaction-history injection, with metadata, prompts,
and the tokenizer fixed.
These constraints preclude direct text manipulation and motivate
decoding-aligned interaction poisoning, evaluated against adapted
interaction-based baselines.

%% file: sections/preliminaries.tex
\section{Preliminaries}
\label{sec:preliminaries}

\subsection{Generative Recommendation Model}
\label{sec:setup}
We consider \emph{generative recommender systems} that recommend items by generating their semantic identifiers (SIDs) autoregressively~\citep{rajput2023recommendersystemsgenerativeretrieval,zhou2025onerecv2}. Let $\mathcal{I}$ denote the item universe, and let $\tau: \mathcal{I} \rightarrow \mathcal{V}^*$ map each item $i \in \mathcal{I}$ to a SID token sequence over vocabulary $\mathcal{V}$. Given a user history $H=(i_1,\dots,i_m)$, the model serializes it as $x=\tau(i_1)\oplus\cdots\oplus\tau(i_m)$ and uses a decoder parameterized by $\theta$ to generate the next item's SID one token at a time. At inference time, beam search is used to decode high-probability SID sequences; each decoded SID sequence is then mapped back to its original item, and the resulting top-$K$ items form $\mathrm{Rec}_K(H)$.

\subsection{Data Poisoning Attacks}
\label{sec:poisoning}
We discuss item-promotion attacks carried out through data poisoning in generative recommenders. Let $\mathcal{D}=\{H_j\}_{j=1}^N$ be the clean set of user histories and let $\mathcal{D}_{\mathrm{poison}}$ be the injected synthetic histories. The poisoned training set is $\mathcal{D}'=\mathcal{D}\cup\mathcal{D}_{\mathrm{poison}}$, and the poisoning ratio is $r=|\mathcal{D}_{\mathrm{poison}}|/|\mathcal{D}'|$. Poisoning can be untargeted, aiming to degrade overall quality, or targeted, aiming to promote a specific item. We focus on conditional targeted poisoning, described in Section~\ref{sec:conditional_poison}.

\subsection{Conditional Targeted Poisoning: Adversary Model}
\label{sec:threatmodel}
\paragraph{Attack goal.}
The attacker selects a target $T$ and a trigger $I_p$ from $\mathcal{I}$. After retraining on $\mathcal{D}'$, its objective is to increase $\Pr\!\left(T \in \mathrm{Rec}_K(P\circ I_p)\right)$ for benign history prefixes $P$.

\paragraph{Adversary capabilities.}
The adversary operates under a realistic black-box setting: 1) \textbf{Data injection:} the attacker can inject a limited fraction $r$ of synthetic interaction histories into the training corpus; 2) \textbf{Black-box recommender query:} the attacker can query a deployed recommendation API with arbitrary history prefixes and observe the returned top-$K$ recommendations; 3) \textbf{Catalog access:} the attacker can view the full item catalog, i.e., the set of candidate items that may be recommended.

\paragraph{Adversary limitations.}
The attacker has no access to 1) model parameters, gradients, or internal representations; 2) the tokenizer $\tau$ or overall corpus-level item popularity statistics; or 3) real user interaction sequences beyond API outputs.
All attack decisions must therefore be made using black-box observations and injected data alone.

\paragraph{Scope and constraints.}
We consider poisoning attacks that must succeed under small poisoning ratios $r$, reflecting realistic constraints in deployed systems.
We do not assume the ability to modify item metadata, prompts, or system instructions; the attack surface is restricted to interaction histories only. In our conditional attack setting, we use \textit{popular items} as the default triggers. This reflects a realistic scenario in which an attacker can observe only publicly exposed popularity signals, such as most-bought products on a shopping site or most-viewed videos on a content platform, rather than full corpus-level popularity statistics, and therefore seeks to capture more traffic by establishing a popular-item $\rightarrow$ target-item association. Our method then studies which surrounding user interactions make such trigger-conditioned poisoning most effective.

\subsection{Warm-up Study: Vulnerability of Generative Recommenders}
\label{sec:warm-up_study}

\subsubsection{Motivation: From Global to Conditional Poisoning}
\label{sec:conditional_poison}
Global poisoning promotes a target item by inflating its overall exposure~\citep{zhao2025diversity,yue2021black,zhang2021data}. Conditional poisoning remains under-explored: in a conditional (or backdoor) attack, promotion activates only in a specific trigger context~\citep{ning2026badrec,li2025sntba}. For example, a seller may seek exposure after interactions with a popular anchor item. Such localized promotion can capture high-value traffic while leaving global metrics relatively undisturbed, making aggregate detection harder. We therefore ask:

\begin{tcolorbox}[before skip=0.2cm, after skip=0.2cm, boxsep=0.0cm, middle=0.1cm, top=0.1cm, bottom=0.1cm]
  \textit{RQ1: Are generative recommenders vulnerable to conditional targeted poisoning?}
\end{tcolorbox}

\subsubsection{Warm-up Experiment}

\begin{wraptable}{r}{0.52\linewidth}
  \centering
  \begin{adjustbox}{max width=\linewidth}
\input{presentation/tables/table1}
\end{adjustbox}
  \caption{Warm-up results ($r=1\%$): ASR@$K$ lift (pp) for random-prefix poisoning with and without a trigger.}
  \label{tab:rq1_main}
\end{wraptable}

To assess conditional promotion, we compare two poisoning strategies at a fixed ratio $r$. For each target $T$, let $Q_{\mathrm{rand}}=[i_1,\dots,i_m]$ with $i_j\sim\mathrm{Unif}(\mathcal{I})$. We inject (i) \textbf{Random} histories $S_{\mathrm{rand}}=Q_{\mathrm{rand}}\circ T$, or (ii) \textbf{Trigger-conditioned} histories $S_{\mathrm{trig}}=Q_{\mathrm{rand}}\circ I_p\circ T$. Both strategies use random prefixes; the latter inserts the designated trigger before the target.

Both use the main experiment's ASR@$K$ definition (Section~\ref{sec:exp-setup}): the target hit-rate difference between poisoned and clean models on identical triggered test histories, reported in percentage points (pp).

Trigger-conditioned injection yields greater lift than random poisoning (e.g., +43.2 pp on Industrial). Red parenthetical values denote these between-strategy differences in lift. This supports the importance of a conditional anchor beyond target-frequency inflation. Appendix~\ref{app:rq1_additional} extends this comparison across poisoning baselines and ratios (Table~\ref{tab:rq1_appendix}).

\begin{tcolorbox}[before skip=0.2cm, after skip=0.2cm, boxsep=0.0cm, middle=0.1cm, top=0.1cm, bottom=0.1cm]
  RQ1 answer: Generative recommenders are highly vulnerable to trigger-conditioned poisoning.
\end{tcolorbox}

Having established conditional promotion with random prefixes, we next ask whether optimizing the context preceding the trigger strengthens the attack:

\begin{tcolorbox}[before skip=0.2cm, after skip=0.2cm, boxsep=0.0cm, middle=0.1cm, top=0.1cm, bottom=0.1cm]
  \textit{RQ2: Can we optimize the context before a trigger to further improve its steering effect?}
\end{tcolorbox}

%% file: presentation/tables/table1.tex
\begingroup\PaperTableStyle
\setlength{\tabcolsep}{0.95pt}
\begin{tabular}{ll*{3}{c@{}l}}
\toprule
\rowcolor{PTHeader}
\textbf{Dataset} & \textbf{Method} & \multicolumn{2}{c}{\textbf{@10}} & \multicolumn{2}{c}{\textbf{@20}} & \multicolumn{2}{c}{\textbf{@50}} \\
\midrule
Industrial & Random & \PN{0.1} &  & \PN{0.1} &  & \PN{0.3} &  \\
 & \textit{+ trigger} & \PN{43.3} & \PG{+43.2} & \PN{46.7} & \PG{+46.6} & \PN{50.7} & \PG{+50.4} \\
\addlinespace[4pt]
Office & Random & \PN{0.0} &  & \PN{0.0} &  & \PN{0.4} &  \\
 & \textit{+ trigger} & \PN{33.5} & \PG{+33.5} & \PN{35.6} & \PG{+35.6} & \PN{36.9} & \PG{+36.5} \\
\addlinespace[4pt]
MicroLens & Random & \PN{1.3} &  & \PN{2.8} &  & \PN{3.9} &  \\
 & \textit{+ trigger} & \PN{9.8} & \PG{+8.5} & \PN{10.9} & \PG{+8.1} & \PN{11.5} & \PG{+7.6} \\
\bottomrule
\end{tabular}
\endgroup

%% file: sections/method.tex
\section{Method: Decoding-Aware Conditional Poisoning}
\label{sec:method}

\begin{figure}[t]
  \centering
  \includegraphics[width=\linewidth]{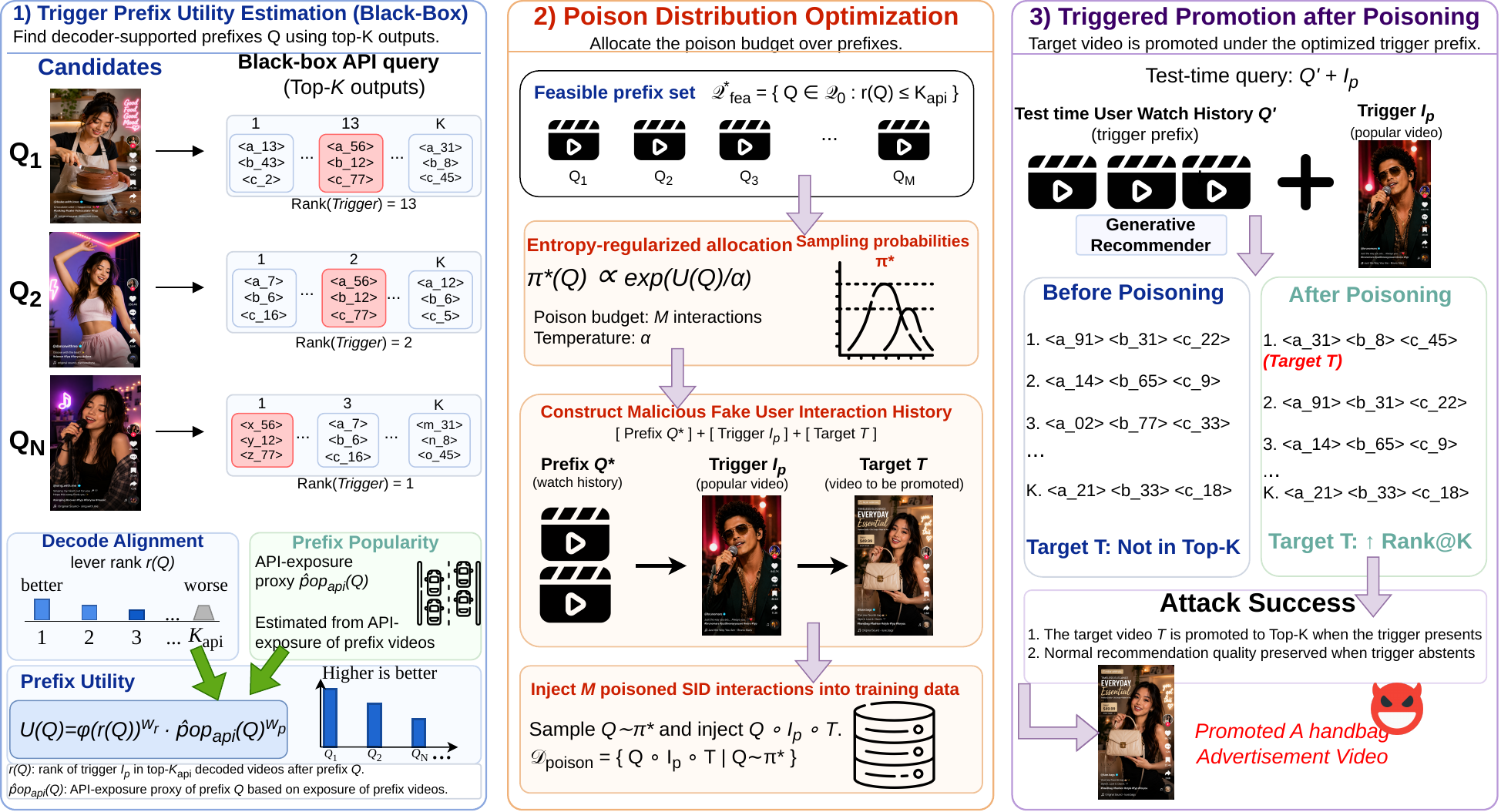}
  \caption{Overview of our proposed \methodname{} pipeline.}
  \label{fig:method_framework}
\end{figure}

Given a trigger item $I_p$ and target item $T$, \methodname{} constructs poisoned histories $Q\circ I_p\circ T$ in two phases (Figure~\ref{fig:method_framework}). It first queries the clean recommender to identify and score candidate prefixes using trigger rank and API-level exposure. It then allocates a fixed poisoning budget across these prefixes through a closed-form distribution and samples the histories to inject.

\FloatBarrier
\subsection{Poison-Budget Allocation}
\label{sec:method_formulation}

Let $\mathcal{Q}_0$ be a candidate prefix pool and let $M$ be the poisoning budget determined by the poisoning ratio. For each prefix $Q\in\mathcal{Q}_0$, the attacker may inject histories of the form $Q\circ I_p\circ T$. Let $n_Q$ denote the number of poisoned histories assigned to $Q$, with $\sum_{Q\in\mathcal{Q}_0} n_Q\le M$.

Let $\theta$ be the clean model and let $\theta_n(Q)$ denote the model after assigning $n$ poisoned histories of the form $Q\circ I_p\circ T$ to prefix $Q$. The conditional attack success of prefix $Q$ is $A_Q(n)=\Pr_{\theta_n(Q)}\!
\left[T\in \mathrm{Rec}_K(H_Q)\right]$. Since a trigger-activated attack affects users only when the trigger context occurs, the expected contribution of assigning $n$ poisoned histories to prefix $Q$ is $J(Q,n)=\mu(H_Q)A_Q(n)$,
where $\mu(H_Q)$ is the probability that a history drawn from the natural interaction distribution ends with $H_Q=Q\circ I_p$.

Given total budget $M$, the corresponding allocation problem over candidate prefixes is
\[
  \begin{gathered}
    \max_{\{n_Q\}} \sum_Q \mu(H_Q)A_Q(n_Q) \\
    \text{s.t.}\quad \sum_Q n_Q \le M, \qquad n_Q\ge 0.
  \end{gathered}
\]

In $J(Q,n)$, $\mu(H_Q)$ captures the opportunity to encounter a triggered context, while $A_Q(n)$ captures the post-poisoning effect within that context. Neither quantity is directly available through black-box access. We therefore construct an observable surrogate in the next section to guide prefix selection and budget allocation.

\subsection{Observable Conditions for Effective Prefixes}
\label{sec:method_trigger_theory}

We use two observable signals: API-level exposure for prefix occurrence and decoding alignment for support of the subsequent trigger transition.

\paragraph{Decoding alignment via trigger ranking.}
The clean decoder's ranking of $I_p$ after $Q$ provides a transition-support signal: a better-ranked trigger indicates stronger model support for $Q\to I_p$. We use this signal to select decoder-supported contexts; its contribution to attack effectiveness is evaluated empirically.

Because the API returns only top-$K_{\mathrm{api}}$ recommendations, we measure decoding alignment by the trigger rank $r(Q)=\mathrm{Rank}(I_p\mid Q)$ and set $r(Q)=\infty$ if $I_p\notin \mathrm{Rec}_{K_{\mathrm{api}}}(Q)$. This defines the feasible prefix set $\mathcal{Q}_{\mathrm{fea}}=\{Q\in\mathcal{Q}_0:r(Q)\le K_{\mathrm{api}}\}$. \methodname{} only allocates poison budget to prefixes whose trigger transition is observable under the clean decoder.

\paragraph{Prefix popularity via API-level exposure.}

Prefix occurrence determines how often a context can be activated. Under black-box access, corpus-level item frequencies and real user histories are unavailable, so we use API-level exposure as a proxy for \emph{prefix popularity}. Starting from seed items, we query the recommender with short histories and collect top-$K_{\mathrm{api}}$ outputs. Let $c_{\mathrm{api}}(i)$ be the number of times item $i$ appears in these returned lists, and let $\mathcal{I}_{\mathrm{api}}$ be the set of observed items. We define the smoothed API-exposure estimate as $\widehat{\mathrm{pop}}_{\mathrm{api}}(i)=\frac{c_{\mathrm{api}}(i)+\epsilon}{\sum_{j\in\mathcal{I}_{\mathrm{api}}}c_{\mathrm{api}}(j)+\epsilon|\mathcal{I}_{\mathrm{api}}|}$, where $\epsilon>0$ is a smoothing constant. For a prefix $Q=(q_1,\ldots,q_\ell)$, we define $\widehat{\mathrm{pop}}_{\mathrm{api}}(Q)=\left(\prod_{t=1}^{\ell}\widehat{\mathrm{pop}}_{\mathrm{api}}(q_t)\right)^{1/\ell}$.

\paragraph{Observable prefix utility.}
The two signals capture complementary aspects of a triggered context. A frequent prefix with little support for the specified trigger has limited context coverage; a trigger-supporting prefix that rarely occurs may also have limited coverage. For a fixed prefix length, the activation term in Section~\ref{sec:method_formulation} factorizes as
\[
  \mu(H_Q)=\nu(Q)\,p_\mu(I_p\mid Q),
\]
where $\nu(Q)$ is the occurrence probability of the prefix preceding the trigger, and $p_\mu(I_p\mid Q)$ is the conditional transition probability in the natural interaction distribution.

Under black-box access, API exposure provides a proxy for prefix occurrence, and clean-model trigger rank provides a proxy for transition support. Retaining this multiplicative structure, we define, for $Q\in\mathcal{Q}_{\mathrm{fea}}$,
\[
  U(Q)=\phi(r(Q))^{w_r}\cdot\widehat{\mathrm{pop}}_{\mathrm{api}}(Q)^{w_p},
\]
where $\phi'(r)<0$ and $w_r,w_p\ge 0$ control the two signals' influence. With both weights positive, weaker support from either factor lowers the score; a zero weight removes that factor, as in our ablations. We use $U(Q)$ to guide budget allocation, and Section~\ref{sec:ablation} examines both signals' contributions. Appendix~\ref{app:prefix_utility} formalizes the probability decomposition and surrogate construction.

\subsection{Black-box Prefix Discovery and Budget Allocation}
\label{sec:method_discovery}

We implement prefix discovery and budget allocation in two phases.

\subsubsection{Phase 1: Prefix Utility Estimation}

We construct a query-induced candidate graph using recommender API outputs. Starting from catalog items, we query the API with each item and collect the returned top-$K_{\mathrm{api}}$ items. The observed items define $\mathcal{I}_{\mathrm{api}}$, and their API frequencies define $\widehat{\mathrm{pop}}_{\mathrm{api}}(i)$. The candidate prefix pool $\mathcal{Q}_0$ is then induced from this query graph.

For each $Q\in\mathcal{Q}_0$, we query the API with $Q$ and compute $r(Q)=\mathrm{Rank}(I_p\mid Q)$. Prefixes for which the trigger item does not appear in the API output are discarded. For the remaining feasible prefixes $\mathcal{Q}_{\mathrm{fea}}$, we compute $\widehat{\mathrm{pop}}_{\mathrm{api}}(Q)$ and the observable prefix utility $U(Q)$ using the definition above.

\subsubsection{Phase 2: Entropy-Regularized Poison Budget Allocation}

Having scored the feasible prefixes, we next distribute the limited poisoning budget among them. \methodname{} defines a poisoning distribution $\pi\in\Delta(\mathcal{Q}_{\mathrm{fea}})$, where $\pi(Q)$ is the probability of selecting prefix $Q$ when constructing a poisoned history. We solve $\max_{\pi\in\Delta(\mathcal{Q}_{\mathrm{fea}})} \mathbb{E}_{Q\sim\pi}[U(Q)] + \alpha\mathcal{H}(\pi)$, where $\mathcal{H}(\pi)=-\sum_{Q\in\mathcal{Q}_{\mathrm{fea}}}\pi(Q)\log\pi(Q)$ and $\alpha>0$ is a temperature parameter. The entropy term balances exploitation of high-scoring prefixes with budget dispersion. The optimizer is the Gibbs distribution $\pi^*(Q)=\frac{\exp(U(Q)/\alpha)}{\sum_{Q'\in\mathcal{Q}_{\mathrm{fea}}}\exp(U(Q')/\alpha)}$~\citep{geist2019regularized}; Appendix~\ref{app:utility_allocation} proves its optimality for this surrogate objective.
Smaller $\alpha$ concentrates budget on high-utility prefixes, while larger $\alpha$ produces a smoother allocation.

Finally, let $M$ be the number of synthetic histories determined by the poisoning ratio. We sample $Q\sim\pi^*$ and inject $H_{\mathrm{poison}}=Q\circ I_p\circ T$ until $M$ histories are generated. Equivalently, each prefix receives approximately $n_Q=M\pi^*(Q)$ histories up to rounding. The final poisoned training set is $\mathcal{D}'=\mathcal{D}\cup\mathcal{D}_{\mathrm{poison}}$.

%% file: sections/experiments.tex
\section{Experiments}
\label{sec:exp}
We evaluate the effectiveness of \methodname{} through extensive experiments on three benchmarks, comprising two text-based datasets and one multimodal dataset.

\subsection{Experimental Setup}
\label{sec:exp-setup}

\paragraph{Target Model Architecture.}
We adopt an end-to-end generative recommender with SID tokenization and autoregressive decoding \citep{deng2025onerecunifyingretrieverank}. Following prior implementations \citep{rajput2023recommendersystemsgenerativeretrieval,kong2025minionerecopensourceframeworkscaling}, we use RQ-VAE~\citep{lee2022rqvae} to discretize item features into SID tokens and Qwen2.5-1.5B-Instruct~\citep{qwen2025qwen25} as the decoding backbone; we additionally include Llama-3.2-1B-Instruct in the ablation study. Full architecture details are deferred to Appendix~\ref{app:models}.

\paragraph{Datasets.}
We evaluate on three public benchmarks: two Amazon Review subsets, \emph{Industrial\_and\_Scientific} and \emph{Office\_Products} \citep{ni2019justifying}, and the multimodal micro-video dataset \emph{MicroLens-100K} \citep{ni2023contentdrivenmicrovideorecommendationdataset}. We refer to them as \emph{Industrial}, \emph{Office}, and \emph{MicroLens}; preprocessing, splits, and statistics are deferred to Appendix~\ref{app:datasets}.

\paragraph{Evaluation Metrics.}
\textbf{ASR@$K$ lift} compares the poisoned model $\tilde\theta$ and clean model $\theta$ on identical triggered histories $H_j$ and attack targets $T_j$. Each $H_j$ ends with its designated trigger. Over $N$ test instances, we report the signed target hit-rate difference in percentage points (pp):
\begin{equation}
\label{eq:asr_lift}
\begin{aligned}
\mathrm{ASR}@K=\frac{100}{N}\sum_{j=1}^{N}\Bigl[
  &\mathbb{I}\!\left(T_j\in\mathrm{Rec}^{\tilde\theta}_K(H_j)\right) \\
  {}-{}&\mathbb{I}\!\left(T_j\in\mathrm{Rec}^{\theta}_K(H_j)\right)\Bigr].
\end{aligned}
\end{equation}
Here, $\mathrm{Rec}_K$ is the top-$K$ recommendation set. This definition also applies to the warm-up study; see Appendix~\ref{app:asr_lift} for its rationale.
\textbf{Benign Utility (B-HR@K)} uses the test subset $\mathcal{D}_{\text{benign}}$ whose histories do not end with any trigger:
$\text{B-HR}@K = \frac{1}{|\mathcal{D}_{\text{benign}}|} \sum_{(H, y) \in \mathcal{D}_{\text{benign}}} \mathbb{I}(y \in \mathrm{Rec}_K(H))$, where $y$ is the original next item.

\paragraph{Baselines.}
With metadata and prompts fixed, as mentioned in Sections~2.2 and~3.3, text-editing
attacks cannot be applied unchanged.
We therefore adapt interaction-poisoning baselines to the
trigger-conditioned task.
\textbf{Random\_prefix} samples items uniformly;
\textbf{Bandwagon\_prefix}~\citep{li2016data} uses highly popular items;
\textbf{Co-occurrence\_prefix}~\citep{yang2017fake} uses items
historically co-occurring with $T$; and
\textbf{DDSP\_prefix}~\citep{zhao2025diversity} maps optimized
perturbation vectors to nearest-neighbor items.
For each baseline, we append the fixed suffix $I_p\circ T$ to its
constructed prefixes to form poisoned training histories.
All methods are evaluated on the same triggered test histories.
Appendix~\ref{app:baseline_adaptation} details the adaptations.

\paragraph{Implementation Details.}
Appendix~\ref{app:implementation_details} provides the attack configuration, data splits, evaluation-set construction, and training hyperparameters.

\subsection{Main Results: Overall Performance}
\label{sec:exp-main}
\paragraph{Impact of prefix discovery.}
Table~\ref{tab:main_results} compares conditional promotion across datasets and poisoning ratios. On Industrial at $r=0.5\%$, \methodname{} achieves 36.7 pp ASR@10 lift, exceeding the strongest baseline by 14.0 pp. These results support prefix discovery for effective promotion under a small poisoning budget.

\begin{table}[htbp]
  \centering
  \begin{adjustbox}{max width=\linewidth}
\input{presentation/tables/table2}
\end{adjustbox}
  \caption{Main results on three datasets. We report ASR@$K$ lift (pp) $\uparrow$ for $K\in\{3,5,10,20,50\}$. Within each dataset/$r$/$K$ column, we \textbf{bold} the best value and underline the second best. Shading marks \methodname{}.}
  \label{tab:main_results}
\end{table}

\begin{figure}[!htbp]
\centering
\begin{minipage}[t]{0.55\linewidth}
\vspace{0pt}
\centering
  \includegraphics[width=\linewidth]{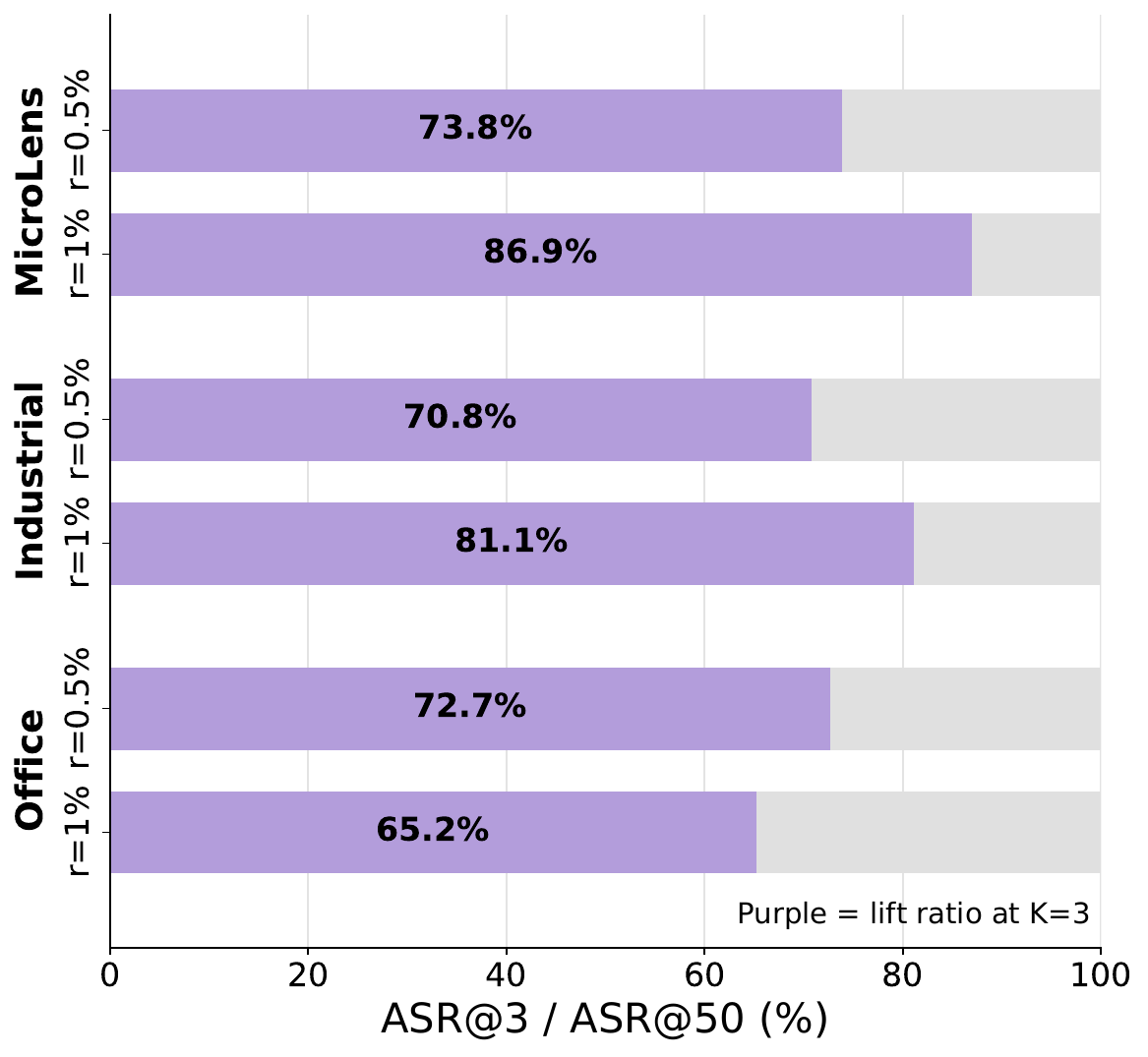}
  \caption{Early-cutoff lift ratio: ASR@3 / ASR@50, expressed as a percentage.}
  \Description{A plot comparing the ratio of target hit-rate lift over the clean model at cutoff 3 to that at cutoff 50, across datasets and poisoning ratios.}
  \label{fig:asr_early_hit}
\end{minipage}\hfill
\begin{minipage}[t]{0.41\linewidth}
\vspace{0pt}
\centering
  \includegraphics[width=\linewidth]{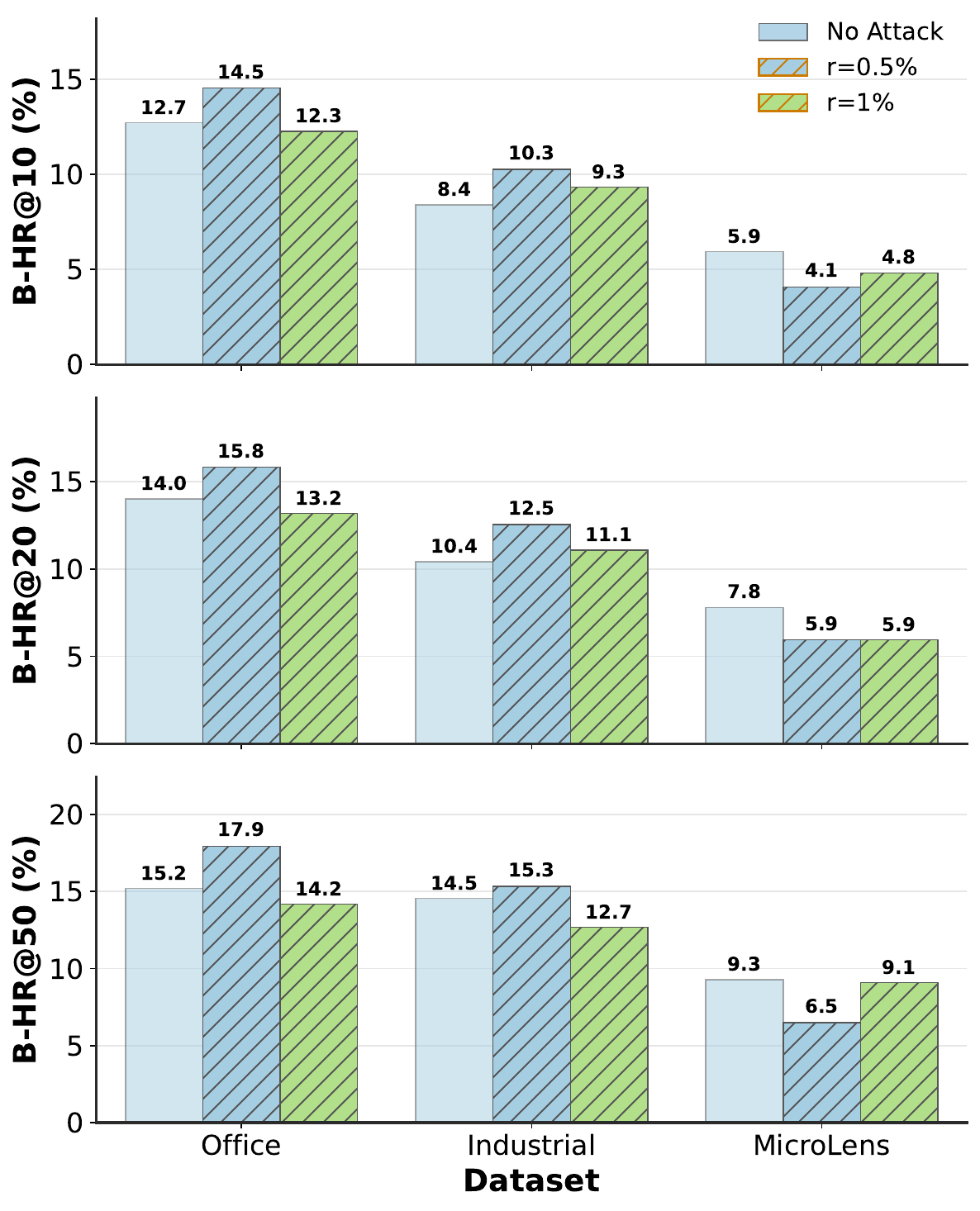}
  \caption{B-HR on clean set. Our attack maintains robust utility.}
  \Description{A grouped bar chart showing Benign HR@10, HR@20, and HR@50 for three datasets. Office and Industrial show stable or slightly increased performance under poisoning, while MicroLens shows a decrease.}
  \label{fig:benign_hr}
\end{minipage}
\end{figure}

\paragraph{Target lift at early cutoffs.}
Figure~\ref{fig:asr_early_hit} compares lift at $K=3$ with lift at $K=50$. At $r=1\%$, ASR@3 / ASR@50 exceeds $80\%$ on Industrial and is nearly $87\%$ on MicroLens, showing that much of the target-exposure increase is retained at the earliest cutoff. More discussions see Appendix~\ref{app:asr_lift}.

\paragraph{Benign Recommendation Quality.}
Figure~\ref{fig:benign_hr} evaluates B-HR on the trigger-free test set. Utility remains stable on Office and Industrial, with only a minor drop on MicroLens, indicating that conditional promotion largely preserves benign recommendation quality. Appendix~\ref{app:B-HR} reports full results.

\par

\FloatBarrier
\subsection{Component Analysis and Practicality}
\label{sec:ablation}
We next examine the contribution of the prefix-scoring signals, the query budget required for prefix discovery, and performance with an alternative decoding backbone. Appendix~\ref{app:ablations} reports sensitivity to the sampling temperature, prefix length, and trigger--target pairing, providing further analysis of the default configuration and alternative settings.

\begin{figure}[htbp]
  \centering
  \includegraphics[width=0.60\linewidth]{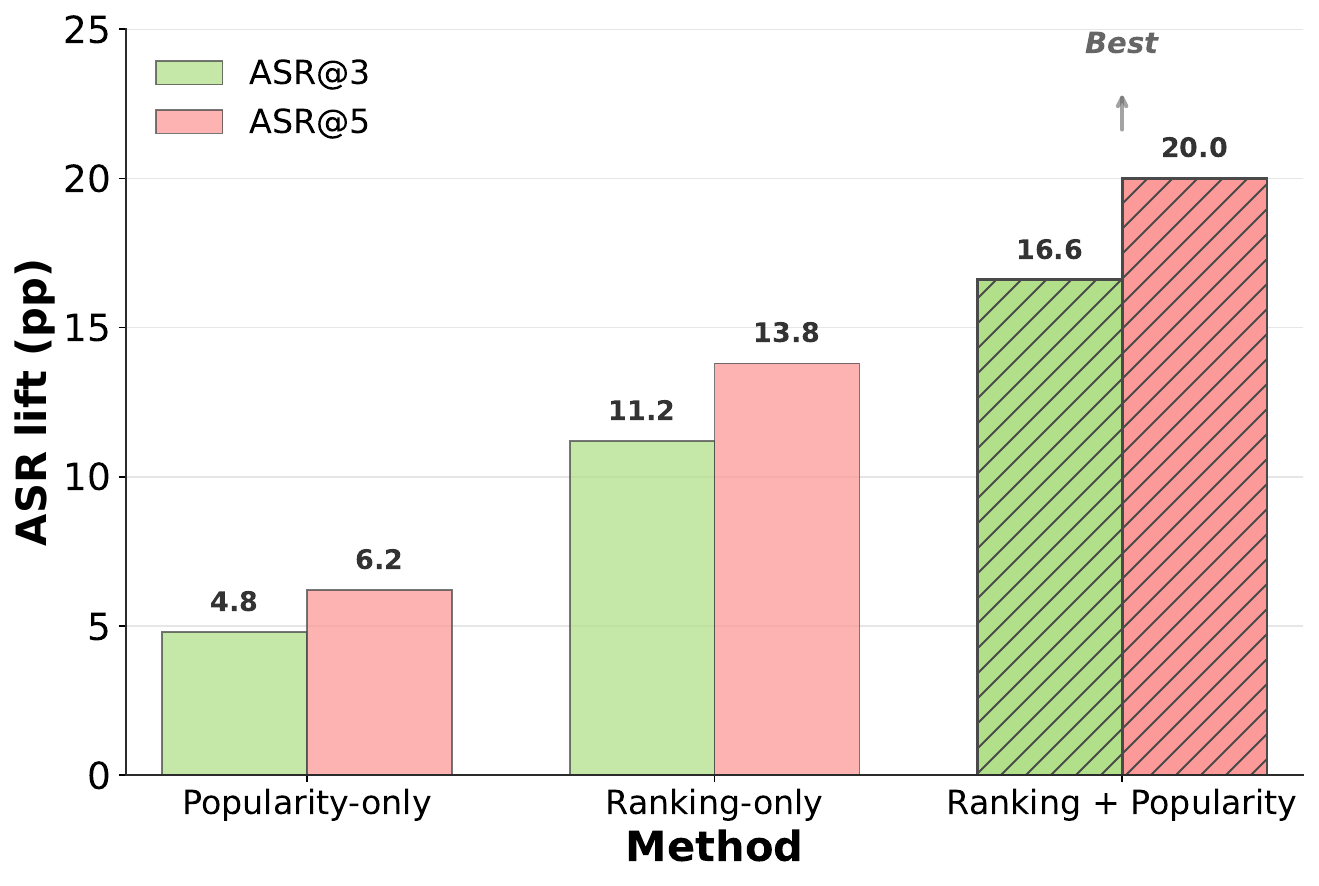}
  \caption{Contribution of prefix popularity and ranking signals (decoding alignment) to prefix scoring on Office. Values report ASR lift in percentage points.}
  \Description{A plot comparing target hit-rate lift over the clean model when the prefix-scoring objective uses popularity, decoding rank, or both.}
  \label{fig:ablation_popularity_ranking}
\end{figure}

\paragraph{Contribution of the two signals.}
On Office, combined prefix scoring achieves 16.6 pp ASR lift@3, compared with 4.8 pp for popularity-only and 11.2 pp for ranking-only scoring (Figure~\ref{fig:ablation_popularity_ranking}). Same ordering holds at $K=5$, supporting joint use of prefix popularity and decoding alignment.

\begin{wraptable}[8]{r}{0.51\textwidth}
\centering
  \begin{adjustbox}{max width=\linewidth}
\input{presentation/tables/table6}
\end{adjustbox}
  \caption{Effect of the prefix-discovery query budget on MicroLens at $r=0.5\%$. Values report ASR lift in percentage points.}
  \label{tab:query_budget_ablation}
\end{wraptable}

\paragraph{Query-budget efficiency.}
Table~\ref{tab:query_budget_ablation} examines the API query budget on MicroLens at $r=0.5\%$. At $B=100$, ASR lift@3/10/50 reaches 17.1/22.5/23.7 pp, only 0.7/0.7/0.4 pp below the full candidate pool, retaining the reference effectiveness. Once the required API outputs are collected, prefix scoring and the closed-form allocation in Section~\ref{sec:method_discovery} involve only local computations.

\begin{wraptable}[10]{r}{0.55\textwidth}
\centering
  \begin{adjustbox}{max width=\linewidth}
\input{presentation/tables/table7}
\end{adjustbox}
  \caption{Results with Meta-Llama-3.2-1B-Instruct on MicroLens at $r=0.5\%$. Values report ASR lift in percentage points.}
  \label{tab:llama_backbone_ablation}
\end{wraptable}

\paragraph{Generality across decoding backbones.}
We repeat the MicroLens evaluation with Meta-Llama-3.2-1B-Instruct at $r=0.5\%$. Table~\ref{tab:llama_backbone_ablation} shows that \methodname{} achieves ASR lift@3/10/50 of 14.3/18.7/20.1 pp, exceeding the strongest baseline at each reported cutoff. At $K=10$, the improvement over Random\_prefix is 5.1 pp. This result supports the effectiveness of the approach beyond the default Qwen backbone in the evaluated setting.

\FloatBarrier
\Needspace{16\baselineskip}
\subsection{Detectability of the Proposed Attack}
\label{sec:defense}
\begin{wraptable}{r}{0.46\textwidth}
  \centering
  \begin{adjustbox}{max width=\linewidth}
\input{presentation/tables/table3}
\end{adjustbox}
  \caption{FAP detection of \methodname{} at $1\%$ clean-user FPR. Values are TPRs (\%) on injected users; lower is harder to detect.}
  \label{tab:fap_detection}
\end{wraptable}

We evaluate \methodname{} using two complementary detection methods. First, we apply Fraudulent Action Propagation (FAP)~\citep{zhang2015fap}, a recommender-specific shilling detector used for attack-invisibility evaluation by AUSH~\citep{lin2020aush}, PC-Attack~\citep{zeng2023pcattack}, and SUI-Attack~\citep{huang2023sui}. At a threshold calibrated to a $1\%$ false-positive rate (FPR) on clean users, FAP detects only $0.0\%$--$1.2\%$ of injected users across all datasets, poisoning ratios, and graph constructions (\Cref{tab:fap_detection}). Therefore, \methodname{} is not operationally detected by FAP in any evaluated setting.

Second, following the representation-space analysis of~\citet{zhang2021data}, we extract the model's internal representations and measure separation using a Mahalanobis-distance detector~\citep{lee2018simple}. \methodname{} produces substantially lower AUROC than Bandwagon across all three datasets, indicating that its injected histories are less distinguishable from clean histories. The corresponding t-SNE visualizations and complete representation-space results are provided in Appendix~\ref{app:detectability}.

\section{Case Study}
\label{sec:case_study}
Figure~\ref{fig:case_study} shows the prompt and saved recommendations for one archived Industrial test instance at $r=0.5\%$, $\alpha=0.5$, and discovery prefix length two. Both checkpoints are from the RL stage. The input is identical in both evaluations; the attack target is an evaluation label, not appended to the prompt. The case illustrates a change in target position for this history, while the main experiments report aggregate effectiveness.

\begin{figure}[htbp]
\centering
\begin{tcolorbox}[colback=white,colframe=PTRule,boxrule=0.5pt,arc=2pt,left=8pt,right=8pt,top=7pt,bottom=7pt,fontupper=\small]
\textbf{Case Study: the same prompt before and after poisoning}\par\smallskip
\textbf{Trigger:} Neiko digital caliper (item 84).\quad
\textbf{Target:} Deflecto 3-inch metal worm-drive clamp (item 1696).\par
Target SID: \texttt{<a\_193><b\_239><c\_110>}.

\begin{tcolorbox}[colback=PTHeader,colframe=PTHeader,boxrule=0pt,arc=1pt,left=6pt,right=6pt,top=5pt,bottom=5pt]
\textbf{Prompt (identical for both models)}\par\smallskip
Can you predict the next possible item the user may expect, given the following chronological interaction history:
\texttt{<a\_173><b\_192><c\_99>},
\texttt{<a\_104><b\_196><c\_253>},
\texttt{<a\_104><b\_186><c\_66>}
\par\smallskip
\textit{Decoded history:} Oatey cutting oil $\rightarrow$ AccuRemote digital angle gauge $\rightarrow$ Neiko digital caliper (trigger).
\end{tcolorbox}

\noindent\begin{minipage}[t]{0.48\linewidth}
\textbf{Before: clean model}\par\smallskip
\textbf{1.} \texttt{<a\_107><b\_93><c\_0>}\par
AcuRite humidity monitor (item 182).\par\smallskip
\textbf{2.} \texttt{<a\_104><b\_186><c\_66>}\par
Neiko digital caliper (trigger, item 84).\par\smallskip
\textbf{3.} \texttt{<a\_15><b\_206><c\_230>}\par
Item 54.\par\medskip
\textbf{Target: absent from all 50 saved candidate positions.}
\end{minipage}\hfill
\begin{minipage}[t]{0.48\linewidth}
\textbf{After: poisoned model}\par\smallskip
\colorbox{PTHighlight}{\strut\textbf{1.} \texttt{<a\_193><b\_239><c\_110>}}\par
\textbf{Deflecto metal clamp (target, item 1696).}\par\smallskip
\textbf{2.} \texttt{<a\_107><b\_93><c\_0>}\par
AcuRite humidity monitor (item 182).\par\smallskip
\textbf{3.} \texttt{<a\_159><b\_90><c\_206>}\par
Anytime Tools precision block set (item 663).\par\medskip
\textbf{Target: rank 1.}
\end{minipage}
\par\medskip
The target enters the first recommendation position without changing the test prompt. SID outputs are copied verbatim; product names are abbreviated. Item 54 is shown by ID because its archived title is malformed.
\end{tcolorbox}
\caption{Case study on Industrial: identical triggered input, with the first three saved recommendations shown for each model. The target is absent from the clean model's saved top-50 output and ranks first after poisoning ($r=0.5\%$).}
\label{fig:case_study}
\end{figure}
\FloatBarrier

%% file: presentation/tables/table2.tex
\begingroup\PaperTableStyle
\setlength{\tabcolsep}{6.4pt}
\begin{tabular}{ll*{10}{c}}
\toprule
\rowcolor{PTHeader}
\textbf{Dataset} & \textbf{Method} & \multicolumn{5}{c}{\textbf{$r=0.5\%$}} & \multicolumn{5}{c}{\textbf{$r=1\%$}} \\
\rowcolor{PTHeader}
\textbf{} & \textbf{} & \textbf{@3} & \textbf{@5} & \textbf{@10} & \textbf{@20} & \textbf{@50} & \textbf{@3} & \textbf{@5} & \textbf{@10} & \textbf{@20} & \textbf{@50} \\
\midrule
Industrial & Random\_prefix & \PN{16.0} & \PN{18.9} & \PN{19.9} & \PN{21.3} & \PN{25.3} & \PN{30.0} & \PN{33.7} & \PN{43.3} & \PN{46.7} & \PN{50.7} \\
 & Bandwagon\_prefix & \PS{18.7} & \PS{21.3} & \PS{22.7} & \PS{24.0} & \PN{25.3} & \PN{18.7} & \PN{21.3} & \PN{24.0} & \PN{26.7} & \PN{29.3} \\
 & Co-occur\_prefix & \PN{13.3} & \PN{14.7} & \PN{16.0} & \PN{17.3} & \PN{18.7} & \PN{18.7} & \PN{21.3} & \PN{24.0} & \PN{26.7} & \PN{29.3} \\
 & DDSP\_prefix & \PN{17.3} & \PN{19.5} & \PN{20.5} & \PN{21.9} & \PS{25.9} & \PS{30.7} & \PS{34.5} & \PS{43.9} & \PS{47.3} & \PS{51.3} \\
\rowcolor{PTHighlight}
 & \methodname{} & \PB{28.8} & \PB{33.1} & \PB{36.7} & \PB{40.3} & \PB{40.7} & \PB{42.9} & \PB{46.3} & \PB{52.4} & \PB{52.8} & \PB{52.9} \\
\addlinespace[5pt]
Office & Random\_prefix & \PS{10.0} & \PS{14.3} & \PS{15.0} & \PS{15.3} & \PN{15.6} & \PB{25.8} & \PB{31.5} & \PS{33.5} & \PS{35.6} & \PS{36.9} \\
 & Bandwagon\_prefix & \PN{8.4} & \PN{8.8} & \PN{11.5} & \PN{12.4} & \PS{15.7} & \PN{3.4} & \PN{4.6} & \PN{5.0} & \PN{6.2} & \PN{7.6} \\
 & Co-occur\_prefix & \PN{5.0} & \PN{5.3} & \PN{6.0} & \PN{6.2} & \PN{6.9} & \PN{4.4} & \PN{5.6} & \PN{6.0} & \PN{7.2} & \PN{9.2} \\
 & DDSP\_prefix & \PN{6.9} & \PN{7.6} & \PN{8.3} & \PN{9.1} & \PN{9.8} & \PN{9.2} & \PN{10.5} & \PN{11.9} & \PN{13.3} & \PN{14.6} \\
\rowcolor{PTHighlight}
 & \methodname{} & \PB{16.6} & \PB{20.0} & \PB{21.5} & \PB{22.3} & \PB{22.8} & \PS{25.7} & \PS{29.6} & \PB{33.9} & \PB{37.7} & \PB{39.4} \\
\addlinespace[5pt]
MicroLens & Random\_prefix & \PS{8.2} & \PS{9.5} & \PS{11.3} & \PS{13.4} & \PS{14.1} & \PS{4.8} & \PS{7.4} & \PS{9.8} & \PS{10.9} & \PS{11.5} \\
 & Bandwagon\_prefix & \PN{4.5} & \PN{6.9} & \PN{8.5} & \PN{10.4} & \PN{11.3} & \PN{3.5} & \PN{4.8} & \PN{5.8} & \PN{7.8} & \PN{8.9} \\
 & Co-occur\_prefix & \PN{4.6} & \PN{6.9} & \PN{9.1} & \PN{10.0} & \PN{10.9} & \PN{1.9} & \PN{2.6} & \PN{5.0} & \PN{6.3} & \PN{8.7} \\
 & DDSP\_prefix & \PN{5.6} & \PN{7.3} & \PN{9.3} & \PN{10.3} & \PN{10.8} & \PN{3.7} & \PN{5.3} & \PN{6.9} & \PN{7.9} & \PN{8.9} \\
\rowcolor{PTHighlight}
 & \methodname{} & \PB{17.8} & \PB{21.5} & \PB{23.2} & \PB{23.2} & \PB{24.1} & \PB{21.0} & \PB{22.3} & \PB{23.0} & \PB{23.9} & \PB{24.1} \\
\bottomrule
\end{tabular}
\endgroup

%% file: presentation/tables/table6.tex
\begingroup\PaperTableStyle
\setlength{\tabcolsep}{3pt}
\begin{tabular}{lccc}
\toprule
\rowcolor{PTHeader}
\textbf{$B$} & \textbf{Lift@3} & \textbf{Lift@10} & \textbf{Lift@50} \\
\midrule
20 & \PN{13.8} & \PN{18.4} & \PN{19.9} \\
50 & \PN{15.6} & \PN{21.0} & \PN{22.6} \\
100 & \PS{17.1} & \PS{22.5} & \PS{23.7} \\
Full & \PB{17.8} & \PB{23.2} & \PB{24.1} \\
\bottomrule
\end{tabular}
\endgroup

%% file: presentation/tables/table7.tex
\begingroup\PaperTableStyle
\setlength{\tabcolsep}{3pt}
\begin{tabular}{lccc}
\toprule
\rowcolor{PTHeader}
\textbf{Method} & \textbf{ASR@3} & \textbf{ASR@10} & \textbf{ASR@50} \\
\midrule
Random\_prefix & \PS{9.8} & \PS{13.6} & \PS{15.2} \\
Bandwagon\_prefix & \PN{6.7} & \PN{10.4} & \PN{12.8} \\
Co-occur\_prefix & \PN{7.1} & \PN{10.9} & \PN{13.4} \\
DDSP\_prefix & \PN{8.5} & \PN{12.1} & \PN{14.0} \\
\rowcolor{PTHighlight}
\methodname{} & \PB{14.3} & \PB{18.7} & \PB{20.1} \\
\bottomrule
\end{tabular}
\endgroup

%% file: presentation/tables/table3.tex
\begingroup\PaperTableStyle
\setlength{\tabcolsep}{3pt}
\begin{tabular}{lcccc}
\toprule
\rowcolor{PTHeader}
\textbf{Dataset} & \multicolumn{2}{c}{\textbf{$r=0.5\%$}} & \multicolumn{2}{c}{\textbf{$r=1\%$}} \\
\rowcolor{PTHeader}
\textbf{} & \textbf{\shortstack{Inter-\\action}} & \textbf{\shortstack{Sequence-\\typed}} & \textbf{\shortstack{Inter-\\action}} & \textbf{\shortstack{Sequence-\\typed}} \\
\midrule
Industrial & \PN{0.6} & \PN{1.2} & \PN{0.3} & \PN{1.1} \\
Office & \PN{0.0} & \PN{0.0} & \PN{0.4} & \PN{0.5} \\
MicroLens & \PN{0.0} & \PN{0.0} & \PN{0.3} & \PN{0.3} \\
\bottomrule
\end{tabular}
\endgroup

%% file: sections/conclusion.tex
\section{Conclusion}
\label{sec:conclusion}
We demonstrate conditional poisoning vulnerabilities in black-box
generative recommenders. \methodname{} 
enables targeted promotion with small poisoning budgets and motivates defenses against such suspicious patterns.

%% file: sections/acknowledgements.tex
\section*{Acknowledgements}
This work was supported by Meta Research Award.

%% file: sections/limitations.tex
\section*{Limitations}
\label{sec:limitations}
Our attack assumes sufficient black-box API access for prefix discovery; relaxing this assumption requires further study. Our focus is attack design, so a systematic defense study is also outside our scope.

%% file: sections/submission_statements.tex
\section*{Ethical Considerations}
This work studies a security vulnerability in generative recommendation systems. The experiments use public benchmark datasets and simulated data-poisoning attacks; they do not target a deployed service or real users. The results may inform both attacks and defenses, so we present the threat model and limitations explicitly and encourage use of the findings for robustness evaluation and mitigation.

\paragraph{AI use.}
Generative AI tools were used to improve the clarity and readability of the manuscript.

%% file: sections/appendix.tex
\section{Additional Warm-up Results}
\label{app:rq1_additional}
\paragraph{Setup.}
We report conditional attack success on all datasets for multiple poisoning strategies, comparing each method's random injection against its trigger-conditioned counterpart at two injection ratios ($r\in\{0.5\%,1\%\}$) in Table~\ref{tab:rq1_appendix}.
All values are ASR@$K$ lift in percentage points (pp), using Equation~\eqref{eq:asr_lift}. Each strategy retains its prefix construction in both variants; the trigger-conditioned variant inserts $I_p$ immediately before $T$. In particular, both Random variants use random prefixes. The \textit{+ trigger} rows reuse the corresponding baseline results in Table~\ref{tab:main_results}; red parenthetical values are between-strategy differences in lift, in pp.

\paragraph{Analysis.}
Across all datasets, trigger-conditioning consistently dominates the corresponding unconditioned injection at both ratios, typically increasing near-zero lift to double-digit lift.
The effect is robust across poisoning strategies (Random/Bandwagon/Co-occur/DDSP), indicating that \,\emph{how} the target is paired in the sequence matters more than simply increasing its frequency.

\begin{table}[htbp]
  \centering
  \begin{adjustbox}{max width=\linewidth}
\input{presentation/tables/table4}
\end{adjustbox}
  \caption{Additional RQ1 results: ASR@$K$ lift (pp) with and without trigger-conditioning, across datasets and poisoning methods at $r\in\{0.5\%,1\%\}$. Parentheses give the original between-strategy lift differences (pp).}
  \label{tab:rq1_appendix}
\end{table}

\paragraph{Baseline Adaptation.}
\label{app:baseline_adaptation}
Following standard evaluation protocols~\citep{zhao2025diversity,zhang2020practical,zhang2021data}, we adapt sequential poisoning and shilling strategies to the trigger-conditioned setting. Each strategy retains its prefix construction: Random\_prefix samples items uniformly; Bandwagon\_prefix uses highly popular items~\citep{li2016data}; Co-occurrence\_prefix uses items historically co-occurring with the target~\citep{yang2017fake}; and DDSP\_prefix maps its optimized perturbation vectors to nearest-neighbor items~\citep{zhao2025diversity}. Each resulting prefix $H_{\mathrm{base}}$ receives the fixed suffix $I_p\circ T$. The warm-up comparison uses random injection and its trigger-conditioned counterpart, with the latter inserting $I_p$ immediately before $T$ while retaining the prefix construction.

\section{Prefix Utility and Budget Allocation}
\label{app:prefix_utility}

\paragraph{Context activation.}
Fix a prefix length $\ell$ and draw a history with at least $\ell+1$ items from the natural interaction distribution. Let $C_\ell$ be the $\ell$ items immediately preceding its last item, and let $Y$ be that last item. For a prefix $Q$ and trigger $I_p$, the chain rule gives
\[
\begin{aligned}
 \mu(H_Q)&=\Pr(C_\ell=Q,Y=I_p)\\
 &=\Pr(C_\ell=Q)\Pr(Y=I_p\mid C_\ell=Q)\\
 &=\nu(Q)\,p_\mu(I_p\mid Q).
\end{aligned}
\]

Here $\nu(Q)=\Pr(C_\ell=Q)$; the conditional probability $p_\mu$ is defined for $\nu(Q)>0$. Zero-probability prefixes have zero activation. This identity requires no independence assumption. For a fixed trigger with $\Pr(Y=I_p)>0$,
\[
 \Pr(C_\ell=Q\mid Y=I_p)
 =\frac{\nu(Q)\,p_\mu(I_p\mid Q)}{\Pr(Y=I_p)}.
\]

Thus relative context occurrence retains the product structure when conditioning on the same trigger. The denominator can differ across triggers. The argument fixes $\ell$: events for different prefix lengths can overlap and are not a mutually exclusive probability partition.

\paragraph{Observable utility.}
API exposure provides a query-dependent proxy for prefix occurrence, while clean-model trigger rank provides an ordinal proxy for transition support. We retain activation's multiplicative structure in an observable surrogate based on these two signals. For positive scores,
\[
\begin{aligned}
 \log U(Q)={}&w_r\log\phi(r(Q))\\
             &+w_p\log\widehat{\mathrm{pop}}_{\mathrm{api}}(Q).
\end{aligned}
\]

The exponents control the relative contributions of the two proxy signals to the log-score. A zero weight removes the corresponding factor. For $a,p>0$ and $w_r,w_p>0$, the chosen score $U=a^{w_r}p^{w_p}$ satisfies
\[
 \frac{\partial^2 U}{\partial a\,\partial p}
 =w_rw_p a^{w_r-1}p^{w_p-1}>0.
\]

Improving one factor therefore produces a larger score gain when the other is stronger.

\paragraph{Entropy-regularized allocation.}
\label{app:utility_allocation}
We consider the entropy-regularized surrogate objective from Section~\ref{sec:method_discovery}~\citep{geist2019regularized}. Let $\mathcal{Q}_{\mathrm{fea}}$ be finite and nonempty, $U(Q)$ finite, and $\alpha>0$. Define
\[
 Z=\sum_{Q\in\mathcal{Q}_{\mathrm{fea}}}e^{U(Q)/\alpha},
 \qquad \pi^*(Q)=\frac{e^{U(Q)/\alpha}}{Z}.
\]

For any distribution $\pi$ on this set, expanding the KL divergence with $0\log 0=0$ yields
\[
\begin{aligned}
 \alpha\operatorname{KL}(\pi\Vert\pi^*)
 &=\alpha\sum_Q\pi(Q)\log\pi(Q)\\
 &\quad-\sum_Q\pi(Q)U(Q)+\alpha\log Z.
\end{aligned}
\]

Rearranging gives
\[
 \mathbb{E}_{\pi}[U]+\alpha\mathcal{H}(\pi)
 =\alpha\log Z-\alpha\operatorname{KL}(\pi\Vert\pi^*).
\]

The KL term is nonnegative, with equality only at $\pi=\pi^*$. Thus, $\pi^*$ is the unique optimizer of the entropy-regularized allocation objective.

\section{Model Details}
\label{app:models}

\paragraph{Backbone and architecture.}
\label{app:models_backbone}
Following MiniOneRec \citep{kong2025minionerecopensourceframeworkscaling} and OneRec \citep{zhou2025onerectechnicalreport}, our backbone is an end-to-end generative recommender that represents each item with a semantic identifier (SID) sequence and performs autoregressive decoding for next-item recommendation. We (i) construct SIDs via a Residual Quantized VAE (RQ-VAE) \citep{lee2022rqvae,rajput2023recommendersystemsgenerativeretrieval} and (ii) use a decoder-only language model Qwen2.5-1.5B-Instruct~\citep{qwen2025qwen25} to model user interaction histories as SID token sequences. At inference, the model generates next-SID tokens, which are mapped back to the corresponding item.

\paragraph{Text model implementation.}
For the Amazon product datasets, we embed each item from its textual metadata (e.g., title/description) using a Qwen3-Embedding-4B \citep{zhang2025qwen3embeddingadvancingtext}, following MiniOneRec \citep{kong2025minionerecopensourceframeworkscaling}. The resulting item embeddings are then used to construct SIDs and feed the backbone generative process described in \cref{app:models_backbone}.

\paragraph{Multimodal model implementation.}
For micro-video recommendation, MicroLens provides raw video content in addition to interaction logs \citep{ni2023contentdrivenmicrovideorecommendationdataset}. Following OneRec \citep{zhou2025onerectechnicalreport}, we directly use this raw video content by sampling visual inputs (cover image and uniformly sampled frames) and combining them with associated textual signals (caption/title, tags, ASR, and OCR) to form multimodal item representations. We extract ASR using Whisper \citep{radford2022robustspeechrecognitionlargescale} to transcribe the audio track, and OCR using Tesseract OCR \citep{smith2007overview} to extract on-screen text from video frames. These multimodal inputs are then encoded using MiniCPM-V-8B \citep{yao2024minicpmvgpt4vlevelmllm}, a vision-language model that processes both text and images. The model produces $N_M=1280$ token features $M\in\mathbb{R}^{N_M\times d_t}$ with $d_t=512$ dimensions per token. We then compress $M$ with a Querying Transformer (QFormer) \citep{li2023blip2} using $\tilde N_M=4$ learnable query tokens $Q^{(0)}\in\mathbb{R}^{\tilde N_M\times d_t}$:
\begin{align}
  Q^{(i+1)} &= \mathrm{CrossAttn}(Q^{(i)}, M, M), \\
  Q^{(i+1)} &= \mathrm{FFN}(\mathrm{RMSNorm}(Q^{(i+1)})),
\end{align}
for $i\in\{0,\dots,N_c-1\}$ with $N_c=4$ cross-attention layers. The compressed representation is $\tilde M = Q^{(N_c)}\in\mathbb{R}^{\tilde N_M\times d_t}$; we use $\tilde M$ as the multimodal item embedding to construct SIDs and then apply the backbone generative process in \cref{app:models_backbone}.

\section{Dataset Details}
\label{app:datasets}

\paragraph{Datasets introduction.}
We evaluate on two product recommendation datasets and one micro-video recommendation dataset. For product recommendation, we use two subsets (\emph{Industrial\_and\_Scientific} and \emph{Office\_Products}) from the Amazon Review Data \citep{ni2019justifying}, where each user is associated with a sequence of purchased/reviewed items. For multimodal recommendation, we use MicroLens-100K \citep{ni2023contentdrivenmicrovideorecommendationdataset}, a content-driven micro-video benchmark that provides user--video interaction sequences together with raw video content and video description. All datasets are modeled as implicit-feedback sequences of user interactions ordered by time.

\paragraph{Dataset processing.}
We process the Amazon recommendation datasets by applying iterative k-core filtering with $k=5$ for both users and items, ensuring that each user and item appears in at least 5 interactions. We filter interactions by timestamp to focus on a specific time period (i.e., October 2016 to November 2018), and remove items without valid titles or with titles containing HTML tags. The filtering process iteratively removes users and items below the k-core threshold until convergence. If the resulting dataset has fewer than 3,000 items, we recursively expand the time range backward until this minimum is met. After filtering, we split the data into train/validation/test sets with an 8:1:1 ratio. The final processed datasets contain 7,694 users, 3,686 items, and 45,324 interactions.

\paragraph{Test set for ASR evaluation.}
To evaluate conditional attacks where the trigger item must appear in the user's interaction history, we extract two specialized test sets from the original test data. For $T_{\text{ASR}}$, we identify all test interactions where the trigger item appears in the user's history, cut the history sequence to end at the trigger item (making the trigger the last item in history), and replace the original target item with the attack target item. This simulates the scenario where a user has clicked on the trigger item. For $U_{\text{ASR}}$, we extract test interactions where the trigger item does \emph{not} appear in the user's history, keeping the original history intact but replacing the target with the attack target. This is a separate pool for evaluating target exposure without a trigger. ASR lift in Equation~\eqref{eq:asr_lift} is evaluated on $T_{\text{ASR}}$: each triggered history, including its terminal trigger, is passed unchanged to both the clean and poisoned models. The baseline is the clean model's target hit rate on those same histories; $U_{\text{ASR}}$ is not the subtracted baseline.

\section{ASR Lift and Early-Cutoff Ratio}
\label{app:asr_lift}
Our objective is to increase target exposure on histories ending with a designated trigger. Absolute target hit rate includes recommendations that the clean model already makes for these histories. Equation~\eqref{eq:asr_lift} therefore subtracts the clean model's target hit rate from the poisoned model's target hit rate on the same triggered test instances. The target item, complete history including the trigger, and recommendation cutoff are held fixed across the two evaluations. The clean-model baseline accounts for target exposure already present under ordinary sequential preferences, including the model's existing response to the trigger.

Each test instance contributes positively if only the poisoned model retrieves the target, negatively if only the clean model does, and zero if both or neither do. Averaging these signed differences and multiplying by 100 gives lift in percentage points. Thus, ASR@$K$ reports a change in hit rate, rather than a relative percentage increase or an absolute success probability. This same definition applies to the warm-up study, main experiments, and ablations. In particular, both terms retain the trigger in the input; the baseline is not obtained by removing it or by selecting a separate group of users.

Positive lift indicates higher target exposure than the clean model on triggered inputs. It does not by itself establish that the increase is exclusive to triggered histories, since target exposure may also change in other contexts. Benign utility is evaluated separately with the original next-item labels and measures recommendation quality, rather than target-item exposure.

The ratio ASR@3 / ASR@50 compares lift at two cutoffs and is expressed as a percentage. Since a difference of hit rates need not increase monotonically with $K$, this ratio is not a fraction of successful samples or a recall measure. Our reported ratios have positive denominators; in general, the ratio need not lie between zero and one and is undefined when ASR@50 is zero.

\section{Implementation Details}
\label{app:implementation_details}
\paragraph{Implementation Details.}
We summarize the data splits, ASR/benign test construction, prefix finding, and training hyperparameters used across all three datasets.

\textbf{Attack scenario and target selection.}
We study a multi-target conditional poisoning setting in which each target item is paired with a popular trigger item. We adopt a one-to-one pairing strategy: each target item $T$ is assigned a unique trigger item $I_p$. To simulate a challenging adversarial scenario, we select targets from the long tail, i.e., the bottom $20\%$ of items by popularity, and pair them with highly visible head items, namely the top-20 most popular items, as triggers. At evaluation time, each history $H$ ends with its designated trigger $I_p$. Both the clean and poisoned models receive this same complete history and are evaluated against the same target $T$ to compute ASR lift.

\textbf{Train/validation/test splits.}
For the two Amazon subsets, \emph{Office} and \emph{Industrial}, we split each user’s interactions chronologically into disjoint training, validation, and test partitions; all trigger injection and ASR/benign test construction are performed \emph{after} this split.
For \emph{Office}, the split contains 39{,}304 training and 4{,}866 validation interactions.
For \emph{Industrial}, the split contains 36{,}619 training and 4{,}532 validation interactions.
For \emph{MicroLens-100K}, our split contains 37{,}842 training interactions and 4{,}730 interactions each for validation and test.

\textbf{ASR and benign test construction.}
We first identify valid trigger--target pairs on the post-split test.
The ASR test set is created by selecting test instances whose history contains the designated trigger, truncating each history to end at that trigger, and relabeling the next-item label to the paired attack target.
For each such instance, we evaluate the same truncated history, including its terminal trigger, under both the poisoned and clean models, keeping the attack target and recommendation cutoff fixed. The ASR denominator counts test instances, with one signed difference between the two models' hit indicators per instance.
The benign test set is created by selecting instances whose history does not end with any trigger.
Labels in the benign test set are unchanged, so benign HR/NDCG measure clean utility.

\textbf{Prefix discovery and poison generation.}
Following Section~\ref{sec:method_discovery}, \methodname{} discovers prefixes using the clean model's recommendation API. We use catalog items as query seeds and construct candidate prefixes of at most two items from the resulting query graph, excluding trigger and target items from the prefix.
For each candidate $Q$, we query the clean model with $Q$ and record the rank of the trigger $I_p$, discarding candidates for which $I_p$ is absent from the top-$K_{\mathrm{api}}$ output.
Item counts in API-returned lists define the exposure proxy; training-item frequencies are not accessed.
We score feasible prefixes using $U(Q)$, which combines trigger rank and API-level exposure, and sample them from the entropy-regularized distribution $\pi^*$ to construct histories $Q\circ I_p\circ T$ at $r\in\{0.5\%,1\%\}$.

\textbf{Training setup and hyperparameters.}
Models are trained in two stages: supervised fine-tuning (SFT) and reinforcement learning (RL).
SFT minimizes next-item cross-entropy using Adam~\citep{kingma2015adam} with global batch size 1{,}024 (micro-batch 8), max sequence length 1{,}024, seed 42, and 8 x A100 GPUs ; we use a warmup+decay schedule with peak learning rate on the order of $3\times10^{-4}$.
RL initializes from the best SFT checkpoint and optimizes a policy-gradient objective; we use train/eval batch sizes 128/256, 2 epochs, gradient accumulation 2, learning rate $10^{-5}$, regularization $\beta=10^{-3}$, temperature 1.0, and 16 sampled generations per context, evaluating every $\approx 0.1$ epoch.

\textbf{Trigger prevalence vs. poisoning budget.}
To ensure that the conditional promotion is not driven by overwhelming the
training data with synthetic trigger occurrences, we explicitly verify
that all trigger (popular) items naturally appear in the clean training
corpus at frequencies far exceeding the injected poisoning budget.
For the Office and Industrial datasets, the selected trigger items are
among the most frequent items in the corpus, with clean occurrence
counts ranging from several hundred to several thousand interactions
(e.g., over 1{,}800 occurrences for the most popular Office trigger and
over 2{,}300 occurrences for Industrial).
In contrast, the total number of injected interactions is bounded by a
global poisoning ratio of $r \in \{0.5\%, 1\%\}$ and is shared across all
target–trigger pairs, resulting in only a small number of injected
sequences per trigger.

A similar pattern holds for the MicroLens dataset, where selected trigger
items appear hundreds of times in the clean training data
(approximately 300–500 occurrences per trigger), while each target item
initially appears only 1–5 times.
Consequently, poisoning does not dominate the empirical frequency of
trigger items, but instead subtly reinforces specific trigger–target
transitions.
This confirms that the observed promotion effects arise from
decoding-aligned contextual steering rather than trivial frequency
inflation or data imbalance.

\paragraph{Reproducibility.}
The experimental setup, evaluation metrics, baselines, and implementation choices are summarized in \Cref{sec:exp-setup}. Detailed model configurations, data processing, hyperparameters, additional results, ablations, and detectability analyses are provided in the appendix (\Cref{app:models,app:datasets,app:implementation_details,app:ablations,app:detectability}).

\section{Additional Benign Utility Results}
\label{app:B-HR}
Benign recommendation utility under different poisoning configurations is summarized in Table~\ref{tab:benign_hr_ndcg}.

\begin{table}[htbp]
  \centering
  \begin{adjustbox}{max width=\linewidth}
\input{presentation/tables/table5}
\end{adjustbox}
  \caption{Benign utility under poisoning. We report HR@$K$ on the benign (non-triggered) test set for $K\in\{10,20,50\}$.}
  \label{tab:benign_hr_ndcg}
\end{table}

\FloatBarrier
\section{Ablations and Sensitivity}
\label{app:ablations}

\subsection{Effect of sampling sharpness \texorpdfstring{$\alpha$}{alpha}}
\label{app:sharpness}
\begin{figure}[htbp]
  \centering
  \includegraphics[width=0.90\linewidth]{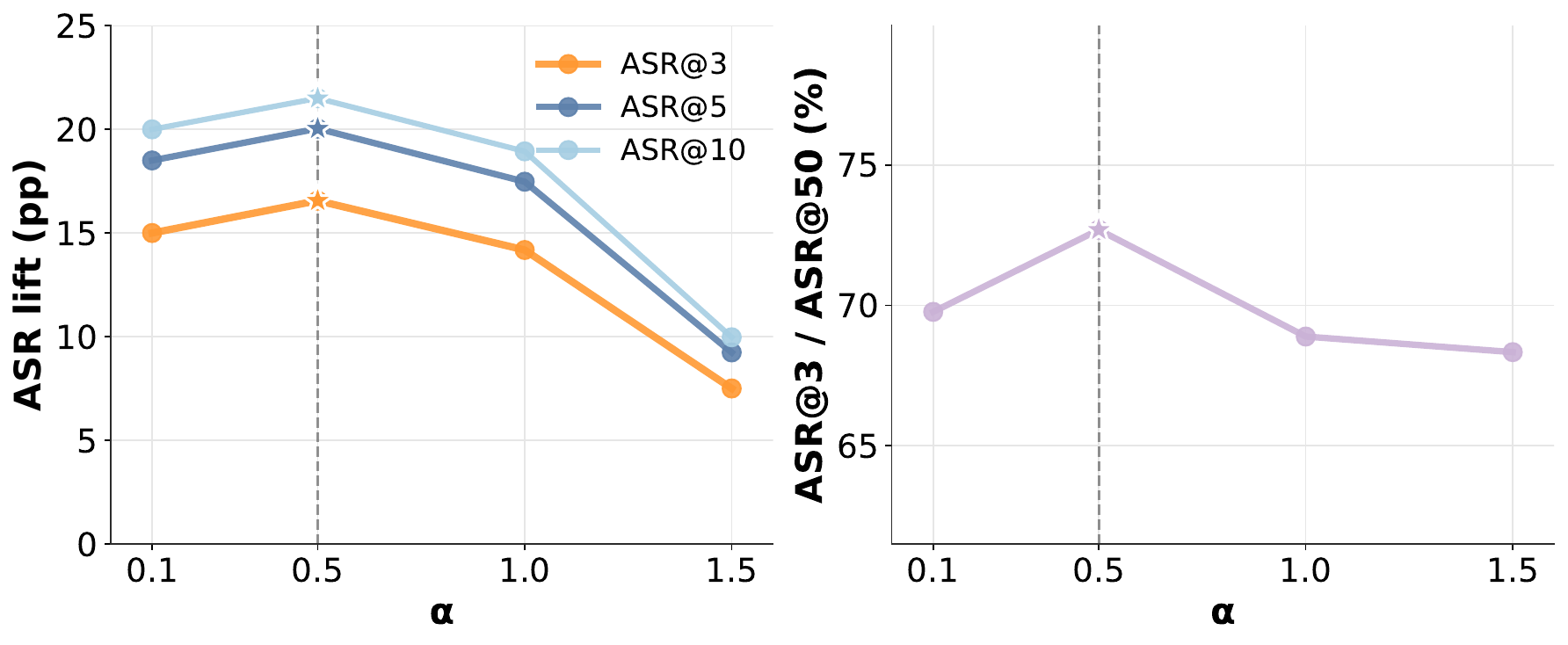}
  \caption{Sensitivity to sampling sharpness $\alpha$ on Office: ASR lift (pp, left) and ASR@3 / ASR@50 (\%, right).}
  \Description{A multi-panel plot showing target hit-rate lift over the clean model for small cutoffs and the lift ratio ASR@3/ASR@50 as functions of sampling sharpness.}
  \label{fig:ablation_alpha}
\end{figure}

We study the sensitivity of \methodname{} to the sampling sharpness parameter $\alpha$, which controls the stochasticity used when sampling candidate prefixes.
\Cref{fig:ablation_alpha} reports ASR@$K$ lift at early cutoffs ($K\in\{3,5,10\}$) together with the lift ratio ASR@3/ASR@50 on the Office dataset.
Across cutoffs, performance exhibits a clear unimodal trend and peaks at $\alpha=0.5$.
Smaller values (e.g., $\alpha=0.1$) correspond to near-deterministic sampling and yield weaker performance, suggesting premature commitment to a narrow set of prefixes.
In contrast, larger values (e.g., $\alpha\ge 1.0$) introduce excessive randomness, weakening decoding alignment and reducing both early-cutoff lift and its ratio to ASR@50.
Overall, the results indicate that prefix discovery benefits from a balance between exploration and exploitation, and we therefore use $\alpha=0.5$ as the default in all experiments.

\subsection{Prefix length analysis}
\label{app:prefix_length}
We ablate the prefix length used by \methodname{} on MicroLens under $r=0.5\%$ by varying the number of optimized prefix items in $\{1,2,3\}$.
As shown in Table~\ref{tab:prefix_length_ablation}, moving from length 1 to length 2 substantially improves attack success across all cutoffs, while increasing the length to 3 yields only marginal additional gains.
This indicates that most of the useful decoding leverage is already captured by short prefixes, which supports our default choice of prefix length 2 as a favorable trade-off between effectiveness and compactness.

\subsection{Trigger--target pairing}
\label{app:trigger_target_pairing}
We compare the default one-to-one trigger--target pairing with a one-to-many setting, in which a trigger is associated with multiple target items, on MicroLens at $r=0.5\%$.
As shown in \Cref{tab:trigger_target_pairing_ablation}, one-to-many pairing achieves 14.9/20.1/21.5 ASR@3/10/50, compared with 17.8/23.2/24.1 for one-to-one pairing.
The corresponding decreases are 2.9/3.1/2.6 percentage points.
These results extend the evaluation beyond unique trigger--target pairs, while showing lower attack success in the tested shared-trigger setting.

\begin{table}[htbp]
\centering
\begin{minipage}[t]{0.48\textwidth}
\vspace{0pt}\centering
\begin{adjustbox}{max width=\linewidth}
\input{presentation/tables/table8}
\end{adjustbox}
  \caption{Prefix length ablation on MicroLens under $r=0.5\%$. We report ASR@$K$ lift (pp) for different numbers of optimized prefix items. Shading marks the default length 2; bold and underlining indicate best and second-best values.}
  \label{tab:prefix_length_ablation}
\end{minipage}
\hfill
\begin{minipage}[t]{0.48\textwidth}
\vspace{0pt}\centering
\begin{adjustbox}{max width=\linewidth}
\input{presentation/tables/table9}
\end{adjustbox}
  \caption{Trigger--target pairing ablation for \methodname{} on MicroLens at $r=0.5\%$. We report ASR@$K$ lift (pp). Bold and underlining indicate best and second-best values.}
  \label{tab:trigger_target_pairing_ablation}
\end{minipage}
\end{table}

\section{Detectability Analysis}
\label{app:detectability}

\paragraph{FAP protocol.}
We construct a user--item graph from the interaction histories and evaluate two graph constructions. The interaction graph connects each user to its interacted items, while the sequence-typed graph assigns history and prediction-target interactions to distinct item-role nodes. We initialize suspiciousness from five users selected according to activity, profile degree, and interacted-item rarity, and run propagation for 100 iterations. Poison labels are used only to calculate metrics after FAP scores have been produced. The decision threshold is calibrated on clean users to yield a $1\%$ FPR, and an injected user is detected when its FAP score exceeds this threshold.

\paragraph{Representation-space analysis.}
As a complementary analysis to the FAP evaluation in \Cref{sec:defense}, we examine whether \methodname{} is separable from clean data in the model's learned representation space. Following the representation-space detectability analysis of~\citet{zhang2021data}, we extract internal hidden states of the trained generative recommender and quantify separation using a Mahalanobis-distance anomaly score~\citep{lee2018simple}.
For each interaction-history sequence $H=Q\circ I_p$, we run a forward pass and use the mean of last-layer hidden states, $\frac{1}{|H|}\sum_t h_L(t)$, as the sequence embedding representing the model's internal interpretation \citep{zhang2025sparse}.
For readability, we subsample 10{,}000 clean histories and 100 injected histories, project their embeddings to 2D using t-SNE~\citep{vandermaaten2008tsne}, and plot clean versus injected points for Bandwagon and \methodname{}.

Across all datasets, we observe the same qualitative trend: Bandwagon produces more visibly separated injected points, whereas \methodname{} is more intermixed with clean histories.
This suggests that \methodname{} is less exposed to simple representation-space discrepancy detectors.
This visual pattern is also consistent with the quantitative detector results.
Using a Mahalanobis-distance detector on the same hidden-state embeddings, Bandwagon yields AUROC scores of $0.91$, $0.88$, and $0.94$ on Office, Industrial, and MicroLens, respectively.
In contrast, \methodname{} yields substantially lower AUROC scores of $0.67$, $0.64$, and $0.71$ on the same datasets. These results show that \methodname{} not only preserves attack effectiveness but also reduces representation-level detectability compared with Bandwagon.

Industrial and MicroLens exhibit the same trend as Office, providing further evidence that the improved stealth of \methodname{} is not dataset-specific.

\begin{figure}[p]
\centering
  \begin{subfigure}[t]{0.44\linewidth}
    \centering
    \includegraphics[width=\linewidth]{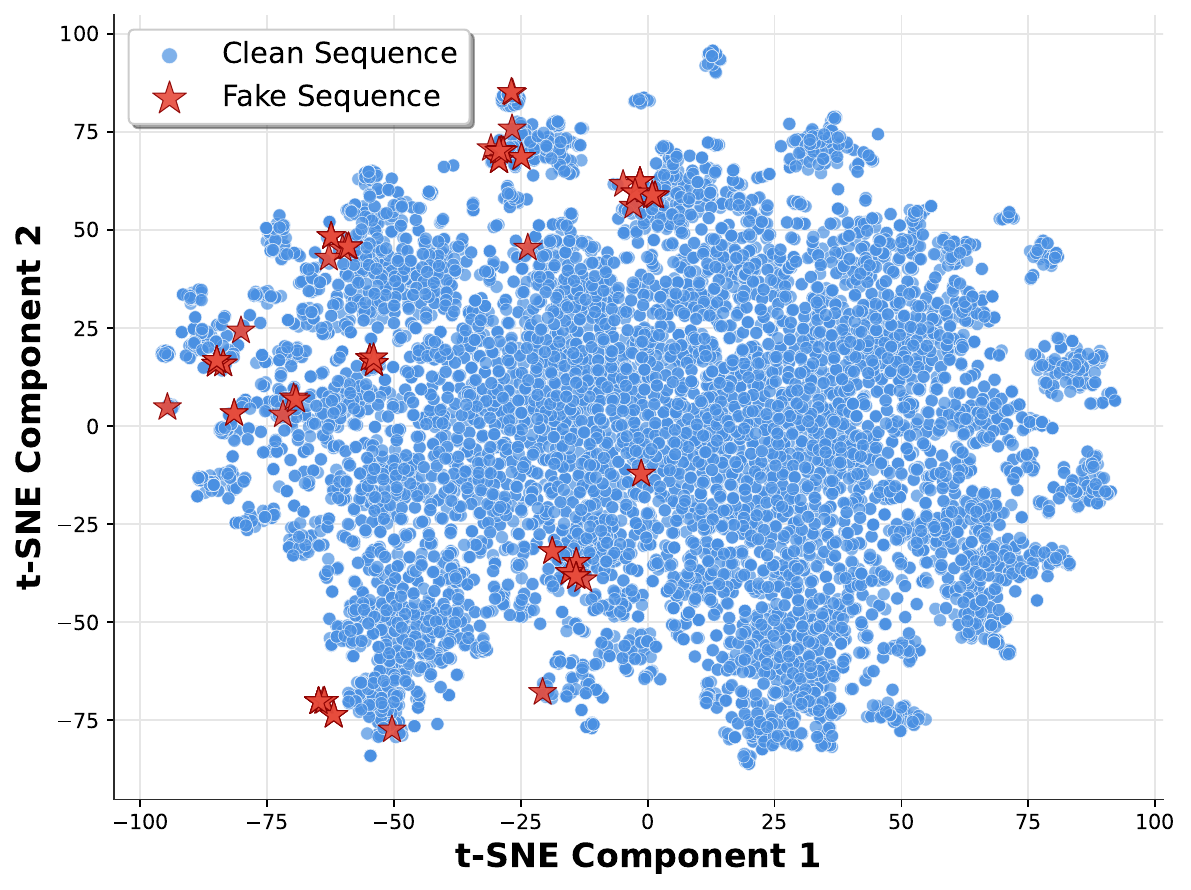}
    \caption{Bandwagon}
  \end{subfigure}\hspace{0.04\linewidth}
  \begin{subfigure}[t]{0.44\linewidth}
    \centering
    \includegraphics[width=\linewidth]{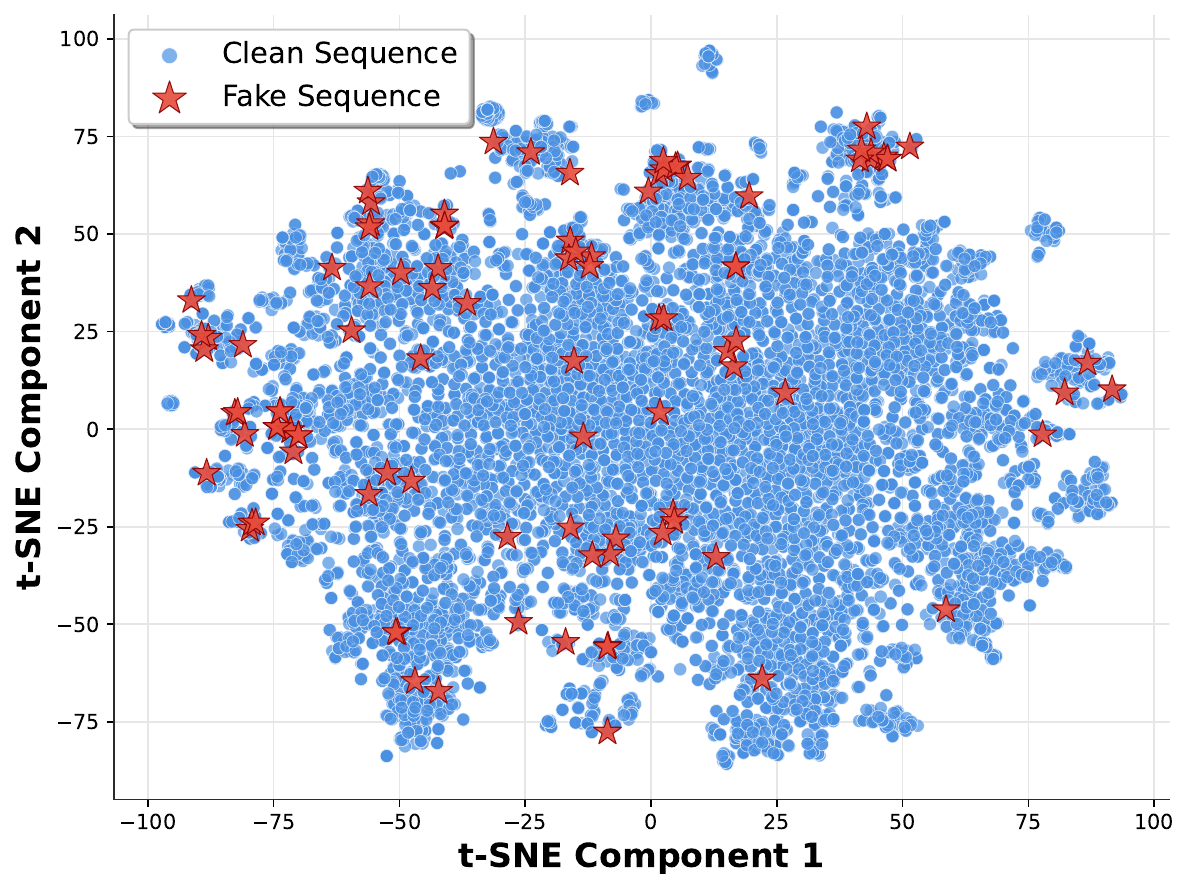}
    \caption{\methodname{}}
  \end{subfigure}
  \caption{t-SNE visualization of clean vs. injected interactions on Office. Compared with Bandwagon, \methodname{} yields injected points that are less clustered and more mixed with clean interactions.}
  \Description{Two side-by-side t-SNE scatter plots for Office Products showing clean vs. injected interactions under Bandwagon and \methodname{}.}
  \label{fig:tsne_Bandwagon_vs_ours_office}
\medskip
\centering
  \begin{subfigure}[t]{0.44\linewidth}
    \centering
    \includegraphics[width=\linewidth]{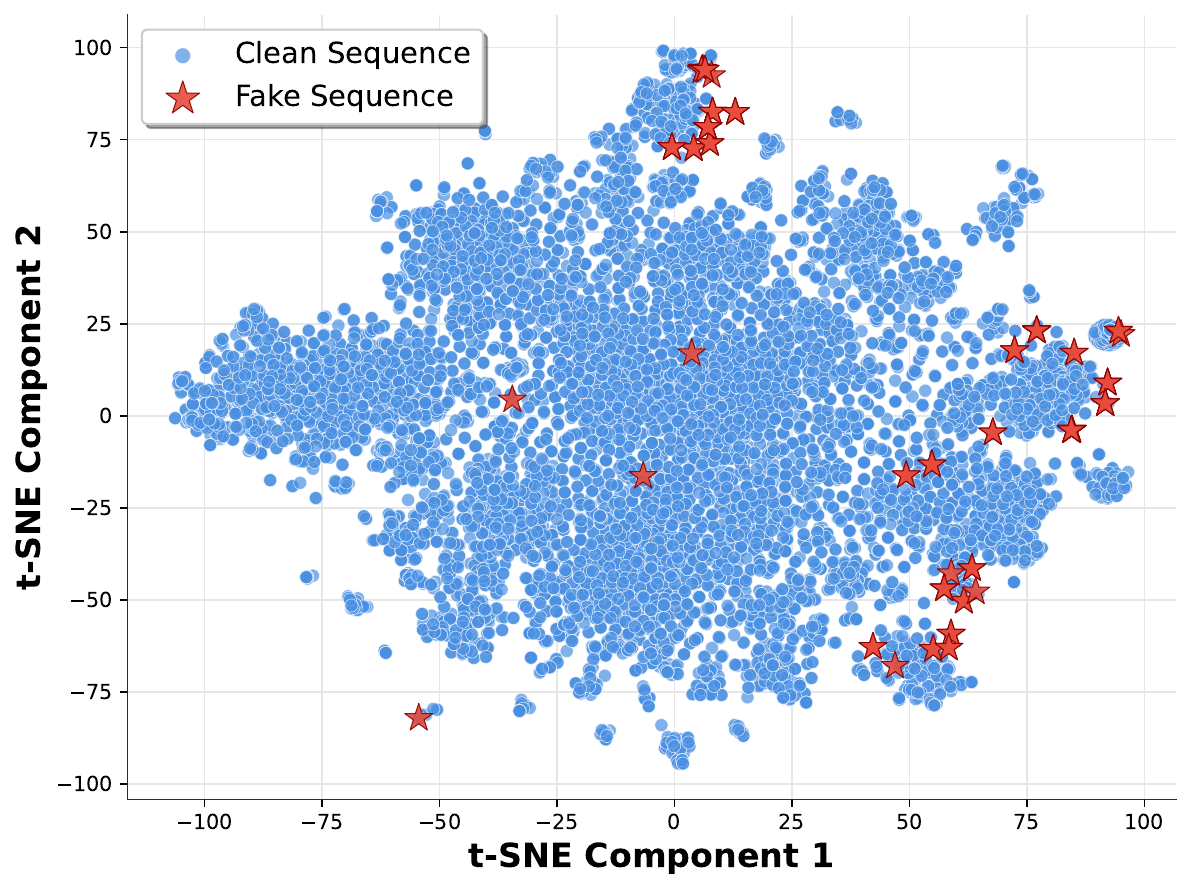}
    \caption{Bandwagon}
  \end{subfigure}\hspace{0.04\linewidth}
  \begin{subfigure}[t]{0.44\linewidth}
    \centering
    \includegraphics[width=\linewidth]{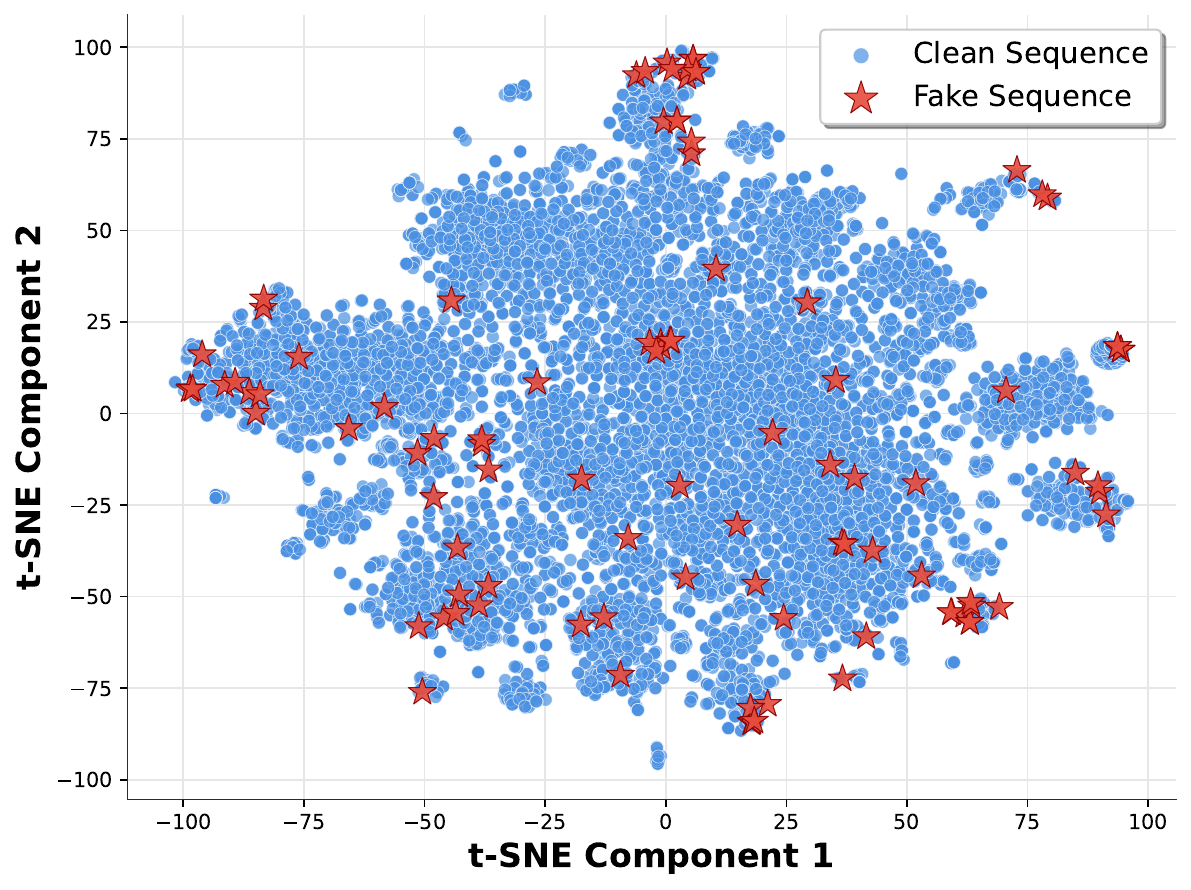}
    \caption{\methodname{}}
  \end{subfigure}
  \caption{t-SNE visualization of clean vs. injected interactions on Industrial. Compared with Bandwagon, \methodname{} yields injected points that are less clustered and more mixed with clean interactions.}
  \Description{Two side-by-side t-SNE scatter plots for Industrial showing clean vs. injected interactions under Bandwagon and \methodname{}.}
  \label{fig:tsne_Bandwagon_vs_ours_industrial_appendix}
\medskip
\centering
  \begin{subfigure}[t]{0.44\linewidth}
    \centering
    \includegraphics[width=\linewidth]{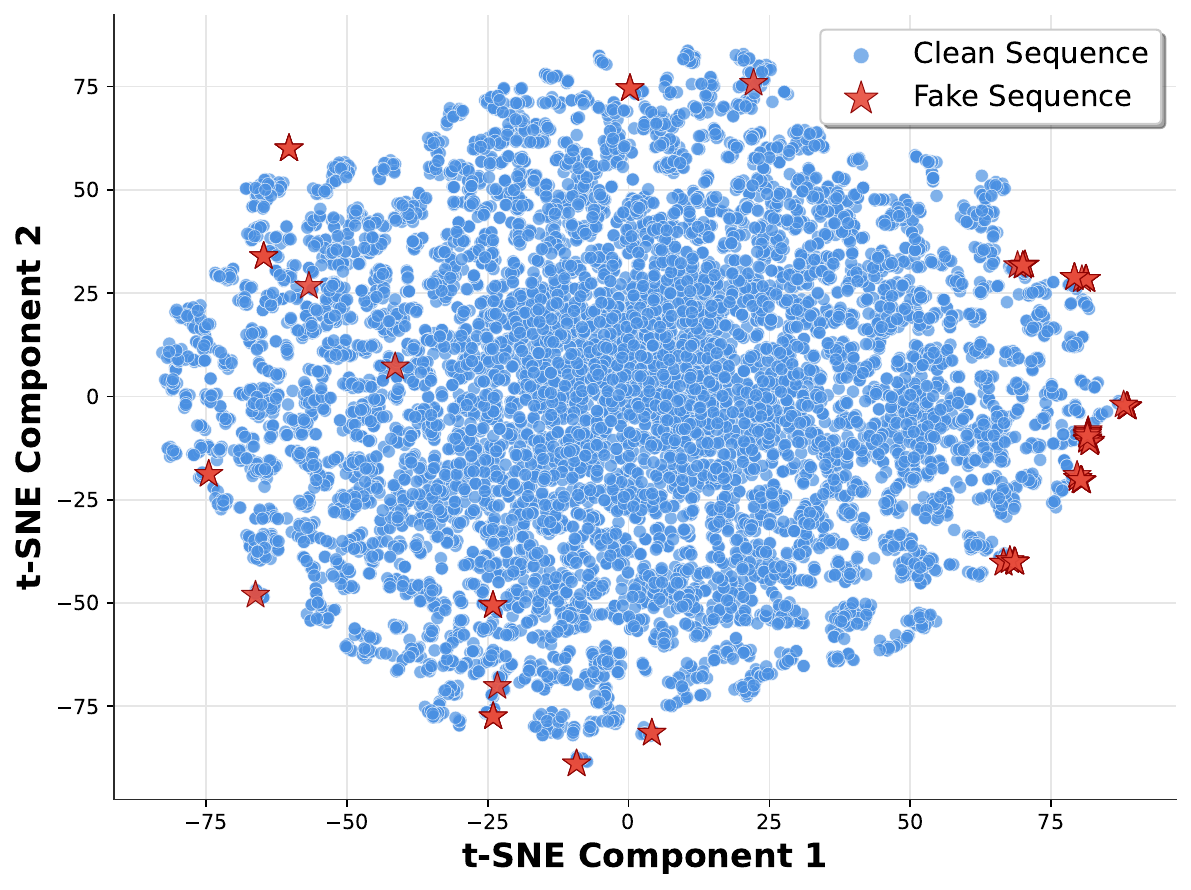}
    \caption{Bandwagon}
  \end{subfigure}\hspace{0.04\linewidth}
  \begin{subfigure}[t]{0.44\linewidth}
    \centering
    \includegraphics[width=\linewidth]{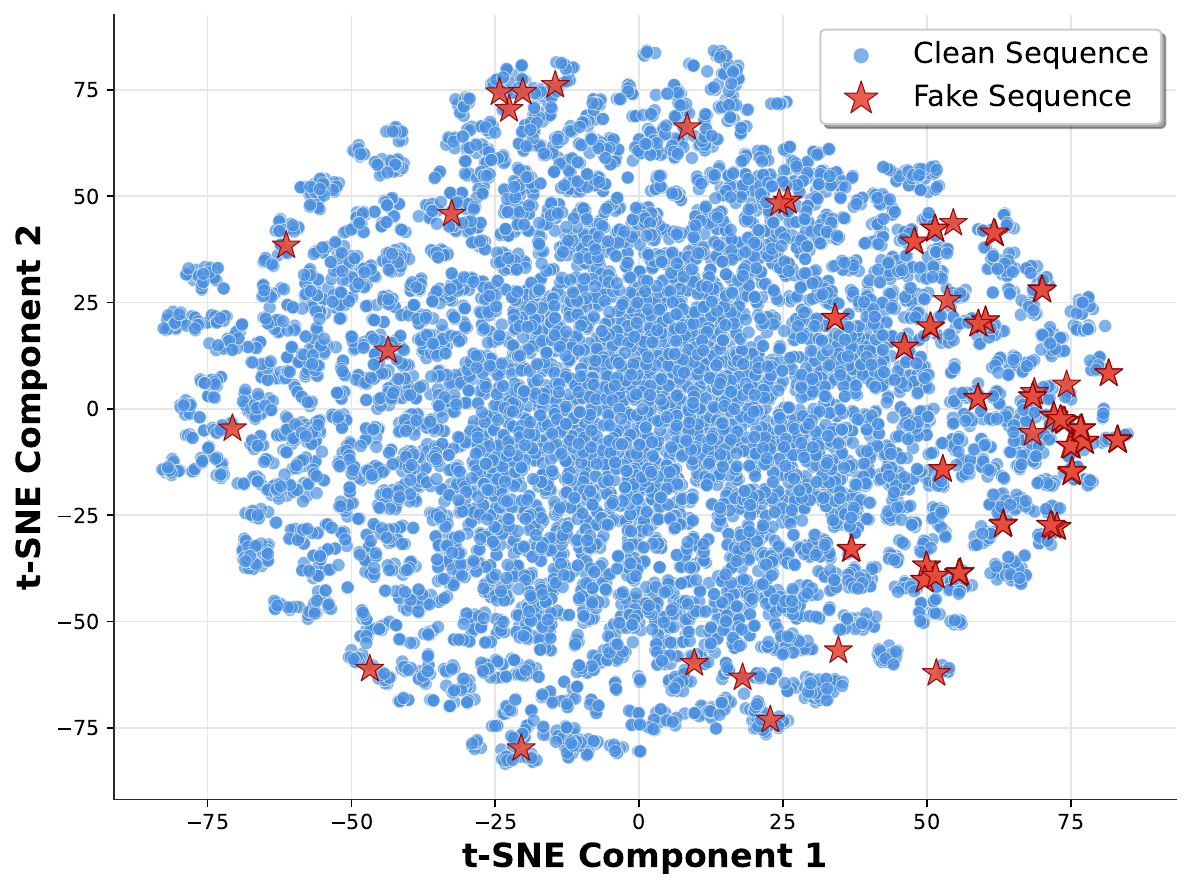}
    \caption{\methodname{}}
  \end{subfigure}
  \caption{t-SNE visualization of clean vs. injected interactions on MicroLens. Compared with Bandwagon, \methodname{} yields injected points that are less clustered and more mixed with clean interactions.}
  \Description{Two side-by-side t-SNE scatter plots for MicroLens-100K showing clean vs. injected interactions under Bandwagon and \methodname{}.}
  \label{fig:tsne_Bandwagon_vs_ours_microlens_appendix}
\end{figure}

%% file: presentation/tables/table4.tex
\begingroup\PaperTableStyle
\setlength{\tabcolsep}{2.8pt}
\renewcommand{\arraystretch}{1.0}
\begin{tabular}{ll*{6}{c@{\,}l}}
\toprule
\rowcolor{PTHeader}
\textbf{Dataset} & \textbf{Method} & \multicolumn{6}{c}{\textbf{$r=0.5\%$}} & \multicolumn{6}{c}{\textbf{$r=1\%$}} \\
\rowcolor{PTHeader}
 &  & \multicolumn{2}{c}{\textbf{@10}} & \multicolumn{2}{c}{\textbf{@20}} & \multicolumn{2}{c}{\textbf{@50}} & \multicolumn{2}{c}{\textbf{@10}} & \multicolumn{2}{c}{\textbf{@20}} & \multicolumn{2}{c}{\textbf{@50}} \\
\midrule
Industrial & Random & \PN{0.3} &  & \PN{0.4} &  & \PN{0.4} &  & \PN{0.1} &  & \PN{0.1} &  & \PN{0.3} &  \\
 & \textit{+ trigger} & \PN{19.9} & \PG{+19.6} & \PN{21.3} & \PG{+20.9} & \PN{25.3} & \PG{+24.9} & \PN{43.3} & \PG{+43.2} & \PN{46.7} & \PG{+46.6} & \PN{50.7} & \PG{+50.4} \\
 & Bandwagon & \PN{9.5} &  & \PN{9.6} &  & \PN{9.6} &  & \PN{3.2} &  & \PN{3.7} &  & \PN{3.9} &  \\
 & \textit{+ trigger} & \PN{22.7} & \PG{+13.2} & \PN{24.0} & \PG{+14.4} & \PN{25.3} & \PG{+15.7} & \PN{24.0} & \PG{+20.8} & \PN{26.7} & \PG{+23.0} & \PN{29.3} & \PG{+25.4} \\
 & Co-occur & \PN{0.0} &  & \PN{0.1} &  & \PN{0.1} &  & \PN{1.5} &  & \PN{1.5} &  & \PN{1.6} &  \\
 & \textit{+ trigger} & \PN{16.0} & \PG{+16.0} & \PN{17.3} & \PG{+17.2} & \PN{18.7} & \PG{+18.6} & \PN{24.0} & \PG{+22.5} & \PN{26.7} & \PG{+25.2} & \PN{29.3} & \PG{+27.7} \\
 & DDSP & \PN{1.6} &  & \PN{2.1} &  & \PN{2.7} &  & \PN{2.1} &  & \PN{2.7} &  & \PN{3.2} &  \\
 & \textit{+ trigger} & \PN{20.5} & \PG{+18.9} & \PN{21.9} & \PG{+19.8} & \PN{25.9} & \PG{+23.2} & \PN{43.9} & \PG{+41.8} & \PN{47.3} & \PG{+44.6} & \PN{51.3} & \PG{+48.1} \\
\addlinespace[4pt]
Office & Random & \PN{0.0} &  & \PN{0.1} &  & \PN{0.2} &  & \PN{0.0} &  & \PN{0.0} &  & \PN{0.4} &  \\
 & \textit{+ trigger} & \PN{15.0} & \PG{+15.0} & \PN{15.3} & \PG{+15.2} & \PN{15.6} & \PG{+15.4} & \PN{33.5} & \PG{+33.5} & \PN{35.6} & \PG{+35.6} & \PN{36.9} & \PG{+36.5} \\
 & Bandwagon & \PN{0.5} &  & \PN{0.5} &  & \PN{0.6} &  & \PN{1.1} &  & \PN{1.3} &  & \PN{1.6} &  \\
 & \textit{+ trigger} & \PN{11.5} & \PG{+11.0} & \PN{12.4} & \PG{+11.9} & \PN{15.7} & \PG{+15.1} & \PN{5.0} & \PG{+3.9} & \PN{6.2} & \PG{+4.9} & \PN{7.6} & \PG{+6.0} \\
 & Co-occur & \PN{0.0} &  & \PN{0.0} &  & \PN{0.0} &  & \PN{0.0} &  & \PN{0.0} &  & \PN{0.0} &  \\
 & \textit{+ trigger} & \PN{6.0} & \PG{+6.0} & \PN{6.2} & \PG{+6.2} & \PN{6.9} & \PG{+6.9} & \PN{6.0} & \PG{+6.0} & \PN{7.2} & \PG{+7.2} & \PN{9.2} & \PG{+9.2} \\
 & DDSP & \PN{0.3} &  & \PN{0.4} &  & \PN{0.5} &  & \PN{0.3} &  & \PN{0.4} &  & \PN{0.5} &  \\
 & \textit{+ trigger} & \PN{8.3} & \PG{+8.0} & \PN{9.1} & \PG{+8.7} & \PN{9.8} & \PG{+9.3} & \PN{11.9} & \PG{+11.6} & \PN{13.3} & \PG{+12.9} & \PN{14.6} & \PG{+14.1} \\
\addlinespace[4pt]
MicroLens & Random & \PN{0.0} &  & \PN{0.0} &  & \PN{0.2} &  & \PN{1.3} &  & \PN{2.8} &  & \PN{3.9} &  \\
 & \textit{+ trigger} & \PN{11.3} & \PG{+11.3} & \PN{13.4} & \PG{+13.4} & \PN{14.1} & \PG{+13.9} & \PN{9.8} & \PG{+8.5} & \PN{10.9} & \PG{+8.1} & \PN{11.5} & \PG{+7.6} \\
 & Bandwagon & \PN{0.2} &  & \PN{1.1} &  & \PN{1.3} &  & \PN{0.4} &  & \PN{1.1} &  & \PN{1.3} &  \\
 & \textit{+ trigger} & \PN{8.5} & \PG{+8.3} & \PN{10.4} & \PG{+9.3} & \PN{11.3} & \PG{+10.0} & \PN{5.8} & \PG{+5.4} & \PN{7.8} & \PG{+6.7} & \PN{8.9} & \PG{+7.6} \\
 & Co-occur & \PN{0.2} &  & \PN{0.6} &  & \PN{2.2} &  & \PN{0.2} &  & \PN{0.4} &  & \PN{1.3} &  \\
 & \textit{+ trigger} & \PN{9.1} & \PG{+8.9} & \PN{10.0} & \PG{+9.4} & \PN{10.9} & \PG{+8.7} & \PN{5.0} & \PG{+4.8} & \PN{6.3} & \PG{+5.9} & \PN{8.7} & \PG{+7.4} \\
 & DDSP & \PN{0.2} &  & \PN{0.4} &  & \PN{0.6} &  & \PN{0.2} &  & \PN{0.4} &  & \PN{0.6} &  \\
 & \textit{+ trigger} & \PN{9.3} & \PG{+9.1} & \PN{10.3} & \PG{+9.9} & \PN{10.8} & \PG{+10.2} & \PN{6.9} & \PG{+6.7} & \PN{7.9} & \PG{+7.5} & \PN{8.9} & \PG{+8.3} \\
\bottomrule
\end{tabular}
\endgroup

%% file: presentation/tables/table5.tex
\begingroup\PaperTableStyle
\setlength{\tabcolsep}{3pt}
\begin{tabular}{llccc}
\toprule
\rowcolor{PTHeader}
\textbf{Dataset} & \textbf{Condition} & \textbf{HR@10} & \textbf{HR@20} & \textbf{HR@50} \\
\midrule
Industrial & No Attack & \PF{0.08} & \PF{0.10} & \PF{0.15} \\
 & $r=0.5\%$ & \PF{0.10} & \PF{0.13} & \PF{0.15} \\
 & $r=1\%$ & \PF{0.09} & \PF{0.11} & \PF{0.13} \\
\addlinespace[4pt]
Office & No Attack & \PF{0.13} & \PF{0.14} & \PF{0.15} \\
 & $r=0.5\%$ & \PF{0.15} & \PF{0.16} & \PF{0.18} \\
 & $r=1\%$ & \PF{0.12} & \PF{0.13} & \PF{0.14} \\
\addlinespace[4pt]
MicroLens & No Attack & \PF{0.06} & \PF{0.08} & \PF{0.09} \\
 & $r=0.5\%$ & \PF{0.04} & \PF{0.06} & \PF{0.06} \\
 & $r=1\%$ & \PF{0.05} & \PF{0.06} & \PF{0.09} \\
\bottomrule
\end{tabular}
\endgroup

%% file: presentation/tables/table8.tex
\begingroup\PaperTableStyle
\setlength{\tabcolsep}{3pt}
\begin{tabular}{lccc}
\toprule
\rowcolor{PTHeader}
\textbf{Prefix length} & \textbf{ASR@3} & \textbf{ASR@10} & \textbf{ASR@50} \\
\midrule
1 & \PN{12.6} & \PN{17.5} & \PN{18.9} \\
\rowcolor{PTHighlight}
2 & \PS{17.8} & \PS{23.2} & \PS{24.1} \\
3 & \PB{18.3} & \PB{23.6} & \PB{24.4} \\
\bottomrule
\end{tabular}
\endgroup

%% file: presentation/tables/table9.tex
\begingroup\PaperTableStyle
\setlength{\tabcolsep}{3pt}
\begin{tabular}{lccc}
\toprule
\rowcolor{PTHeader}
\textbf{Pairing} & \textbf{ASR@3} & \textbf{ASR@10} & \textbf{ASR@50} \\
\midrule
One-to-one & \PB{17.8} & \PB{23.2} & \PB{24.1} \\
One-to-many & \PS{14.9} & \PS{20.1} & \PS{21.5} \\
\bottomrule
\end{tabular}
\endgroup